\documentclass[11pt]{article}

\usepackage[preprint]{acl}

\usepackage{amsmath}
\usepackage{amsfonts}   % for \mathbb
\usepackage{amssymb}
\usepackage{times}
\usepackage{algorithm}
\usepackage{algpseudocode}
\usepackage{latexsym}
\usepackage{multirow}
\usepackage{arydshln}
\usepackage{placeins}
\usepackage[T1]{fontenc}
\usepackage[utf8]{inputenc}

\newcommand{\Input}{\State \textbf{Input: }}
\newcommand{\Output}{\State \textbf{Output: }}
\usepackage{microtype}

\usepackage{inconsolata}

\usepackage{graphicx}
\usepackage{dblfloatfix}
\usepackage{booktabs}
\usepackage{arydshln}
\usepackage[most]{tcolorbox}
\usepackage[dvipsnames]{xcolor}

\title{CLewR: Curriculum Learning with Restarts\\ for Machine Translation Preference Learning}

\author{
Alexandra Dragomir$^{1}$, Florin Brad$^{1}$, Radu Tudor Ionescu$^{2,\diamond}$\\
$^1$Bitdefender, Bucharest, Romania\\
$^2$Department of Computer Science, University of Bucharest, Bucharest, Romania\\
$^{\diamond}$\texttt{raducu.ionescu@gmail.com}
}

\begin{document}

\maketitle
\begin{abstract}
\vspace{-0.2cm}
% draft
Large language models (LLMs) have demonstrated competitive performance in zero-shot multilingual machine translation (MT). Some follow-up works further improved MT performance via preference optimization, but they leave a key aspect largely underexplored: the order in which data samples are given during training. We address this topic by integrating curriculum learning into various state-of-the-art preference optimization algorithms to boost MT performance. We introduce a novel \textbf{c}urriculum \textbf{le}arning strategy \textbf{w}ith \textbf{r}estarts (CLewR), which reiterates easy-to-hard curriculum multiple times during training to effectively mitigate the catastrophic forgetting of easy examples. We demonstrate consistent gains across several model families (Gemma2, Qwen2.5, Llama3.1) and preference optimization techniques. We publicly release our code at  %\url{https://link.hidden.4.blind.review}.
\url{https://github.com/alexandra-dragomir/CLewR}.
\end{abstract}

\setlength{\abovedisplayskip}{3.5pt}
\setlength{\belowdisplayskip}{3.5pt}

\vspace{-0.1cm}
\section{Introduction}
\vspace{-0.1cm}

Large language models (LLMs) have enabled zero-shot approaches in multilingual machine translation (MT) \cite{touvron2023llama}. Methods for improving the MT abilities of LLMs can be broadly categorized into pre-training and post-training approaches. The former typically employ continual pre-training over large-scale monolingual or high-quality parallel data \citep{alves2024tower,cui2025multilingual,xu2023paradigm}.
In contrast, post-training approaches aim to improve translation quality by employing preference optimization techniques, such as Direct Preference Optimization (DPO) \citep{rafailov2023direct}, to distinguish high-quality translations from low-quality ones. Building on this line of work, \citet{xu2024contrastive} proposed Contrastive Preference Optimization (CPO), a reference-free technique that assesses pair distances based on log-probability differences only. More recently, \citet{xu2024x} introduced Adaptive Rejection Preference Optimization (ARPO), which further improves CPO by incorporating an adaptive penalty for the unpreferred term.

Despite the significant advances in preference optimization (PO) techniques \cite{rafailov2023direct,xu2024x, xu2024contrastive}, a key factor that can significantly influence performance remains underexplored: the order in which data samples are processed during training. This aspect is central to \emph{curriculum learning} \citep{bengio2009curriculum}, a paradigm that studies how models can learn from easy to hard. The survey of \citet{Soviany-IJCV-2022} explains that an easy-to-hard learning can be created by manipulating distinct factors, namely the data \cite{Chang-EACL-2021,Jarca-ECAI-2024,Nagatsuka-NGC-2023}, the model \cite{Croitoru-IJCV-2025,Sinha-NeurIPS-2020} or the target task \cite{Liu-IJCAI-2020,Narvekar-AAMAS-2016}. Organizing the samples in a certain order falls under the umbrella of \emph{data-level curriculum}. In the realm of data-level curriculum, researchers have explored both easy-to-hard and hard-to-easy data organizations, the latter being known as anti-curriculum \cite{Ankner-EACL-2024,Florensa-CoRL-2017,Jarca-ACL-2025}. Regardless of the data organization, several recent studies showed that curriculum learning can play an important role in various tasks, e.g.~natural language inference \cite{Poesina-ACL-2024}, intent detection \cite{Gong-CIKM-2021}, question answering \cite{Liu-IJCAI-2018}, image classification \cite{Liu-CVPR-2022}, model pre-training \cite{Madan-WACV-2024,Nagatsuka-NGC-2023}, etc. Curriculum learning has also been applied in neural machine translation (NMT) \cite{Kocmi-RANLP-2017,Liu-ACL-2020,Platanios-NAACL-2019,Zhan-AAAI-2021}, but contributions in this area predate the era of LLMs, making them hard or impossible to adapt to the novel ``pre-training then fine-tuning'' paradigm. With the emergence of preference optimization techniques applied during the fine-tuning stage \cite{rafailov2023direct,xu2024x, xu2024contrastive}, some recent works \cite{Croitoru-CVPR-2025,pattnaik2024enhancing} have integrated curriculum learning into DPO. However, such techniques do not explicitly address catastrophic forgetting \cite{Kirkpatrick-PNAS-2017}, a problem that occurs when samples learned at the beginning are forgotten by the model by the end of the training process, eventually degrading performance.

To this end, we propose a novel data-level curriculum learning framework for MT, in which the easy-to-hard training is restarted at every epoch. Our \textbf{c}urriculum \textbf{le}arning strategy \textbf{w}ith \textbf{r}estarts (CLewR) is natively designed to mitigate catastrophic forgetting by iterating through all samples in every training epoch. We empirically demonstrate that CLewR leads to consistent performance gains in MT across several state-of-the-art preference optimization methods (DPO, CPO, ARPO) and LLM families (Gemma2, Qwen2.5, Llama3.1). Our results show that CLewR not only enhances highly competitive preference optimization methods, but also surpasses another competitor based on curriculum learning, namely CurriDPO \cite{pattnaik2024enhancing}.

In summary, our contribution is threefold:
\begin{itemize}
    \item \vspace{-0.15cm} We propose curriculum learning with restarts (CLewR), a novel method for preference optimization in MT, where the easy-to-hard curriculum is restarted at every epoch to avoid catastrophic forgetting.
    \item \vspace{-0.2cm} While previous work enhanced DPO with curriculum \cite{pattnaik2024enhancing}, we introduce curriculum learning to newer preference optimization algorithms, namely CPO and ARPO.
    \item \vspace{-0.2cm} We show that our method outperforms competing curriculum approaches and consistently improves performance across multiple model families (Gemma2, Llama3.1, Qwen2.5) and preference optimization algorithms (DPO, CPO, ARPO).
\end{itemize}

\vspace{-0.1cm}
\section{Method}
\vspace{-0.1cm}

% \textcolor{red}{TODO: We finetune LoRA adapters on top of the base models and use rank=$64$, learning rate=$5e-5$, warmup ratio=$0.1$, batch size=$4$ .}

%BERTScore \citep{zhang2019bertscore}
%We adopt a data-level curriculum strategy, which we name CLewR, in which the training preference triplets ($x$, $y_{w}$, $y_{l}$) are ordered based on a similarity score $s(y_{w}$, $y_{l})$ computed between the chosen $y_{w}$ and rejected $y_{l}$ responses. 
%Specifically, we define $s$ as the average of normalized BLEU~\citep{papineni2002bleu}, METEOR~\citep{banerjee2005meteor}, and COMET-22~\citep{rei2022comet} scores. Curriculum training starts from easier preference triplets (low similarity) to more difficult (high similarity) for each epoch.  \textcolor{red}{TODO: this part needs to be expand; it is the main contribution; mention catastrophic forgetting}

\noindent\textbf{CLewR.}
We propose a data-level curriculum strategy, named CLewR, which is tailored to preference optimization in MT. We formally present our curriculum strategy in Algorithm~\ref{alg:clewr}. Training preference triplets of the form ($x$, $y_{w}$, $y_{l}$) are ordered (in step 9) based on a similarity score $s(y_{w}$, $y_{l})$ between the chosen (winning) $y_{w}$ and rejected (losing) $y_{l}$ translations. More precisely, the easiness of a pair of translations is defined as the similarity difference between the preferred and rejected translation, i.e.~a high difference corresponds to an easy pair and a low difference to a hard pair. The similarity score is given by the average (step 7) of multiple MT metrics, thus making CLewR suitable for translation: BLEU~\citep{papineni2002bleu} (step 3), COMET-22~\citep{rei2022comet} (step 4), and METEOR~\citep{banerjee2005meteor} (step 5) scores. 

We emphasize that our method implicitly works with multiple correct reference translations for a given source sentence. By default, preference optimization techniques work with triplets of the form ($x$, $y_{w}$, $y_{l}$). If the dataset includes $k$ preferred outputs for the same input, we can build $k$ preference optimization triplets. Then, CLewR can simply apply PO starting from the easier tuples (references that are most dissimilar to the rejected sample) to the more difficult tuples.
\begin{algorithm}[t]
\caption{CLewR Preference Optimization}
\label{alg:clewr}
\begin{algorithmic}[1]
\Input initial policy $\pi_{\theta}$, training triplets $\{(x_i, y_i^{w}, y_i^{l})\}_{i=1}^{N}$, learning rate $\mu$. 

\For{$i = 1$ to $N$}
    \State $b \leftarrow \texttt{BLEU}(y_i^{w}, y_i^{l})$
    \State $c \leftarrow \texttt{COMET}(y_i^{w}, y_i^{l})$
    \State $m \leftarrow \texttt{METEOR}(y_i^{w}, y_i^{l})$
    \State $\hat{b}, \hat{c}, \hat{m} \leftarrow \texttt{normalize}_{(0,1)}(b,c,m)$
    \State $s_i \leftarrow \frac{1}{3} (\hat{b}+\hat{c}+\hat{m})$
\EndFor

\State $\mathcal{I} \leftarrow \texttt{argsort}_{\uparrow}\left(\{s_i\}_{i=1}^{N} \right)$

% \Comment{Lower $s_i$ indicates an easier example}

\For{epoch $= 1$ to $E$}
    \ForAll{batches $\mathcal{B} \subset \mathcal{I}$}
        \State $\mathcal{L}_{\text{PO}} \leftarrow
        \texttt{loss}\!\left(
        x_{\mathcal{B}},
        y^{w}_{\mathcal{B}},
        y^{l}_{\mathcal{B}}, \pi_\theta
        \right)$
        %\State $\theta \leftarrow \theta - \mu \cdot \nabla_{\theta} \mathcal{L}_{\text{PO}}$
        \State $\theta \leftarrow \texttt{optimize}(\theta, \mu, \nabla_{\theta} \mathcal{L}_{\text{PO}})$

    \EndFor
\EndFor
\Output optimized model $\pi_{\theta}$
\end{algorithmic}
\end{algorithm}

We provide examples of easy and hard preference samples in Appendix~\ref{sec:examples}.
After sorting the triplets, training proceeds over a number of epochs (steps 10-15). At every epoch, the samples are divided into mini-batches (step 11) in the exact order established at step 9, i.e.~there is no random shuffling involved. The easy-to-hard data permutation is reused at every epoch, which helps mitigate catastrophic forgetting. Learning is performed via a given PO method (steps 12-13). Note that fixing the order of samples in each epoch does not imply overfitting the order, i.e.~the empirical risk does not depend on sample order. On the contrary, curriculum learning theory \citep{bengio2009curriculum} suggests that organizing the samples in a meaningful order can improve training dynamics, potentially leading to faster convergence and/or better optima.

%This is achieved by introducing several model-based scores for organizing the data samples

\noindent\textbf{CLewR-$\mathbf{z}$.}
For ARPO \citep{xu2024x}, we develop an alternative CLewR implementation called CLewR-$z$, where the curriculum score $s$ (used in step 7 of Algorithm \ref{alg:clewr}) is derived from the ARPO distance $z$. The ARPO objective modifies the CPO objective to incorporate an adaptive penalty term $\tau_{\theta}$, which controls the importance of the rejected term $y_l$:
\begin{equation}
\begin{aligned}
\mathcal{L}_{\text{ARPO}}
\!=\! {} & -\mathbb{E}_{(x,y_w,y_l) \sim \mathcal{D}} \Big[
      \log \sigma \Big(
        \beta \log \pi_{\theta}(y_w \!\mid\! x) \\
&  - \tau_{\theta}(y_w, y_l)\,
          \beta \log \pi_{\theta}(y_l \!\mid\! x)
      \Big) \\
&  + \log \pi_{\theta}(y_w \!\mid\! x)
    \Big].
\end{aligned}
\end{equation}
The value of $\tau_{\theta}$ measures the similarity between $y_w$ and $y_l$, ranging from $0$ to $1$:
\begin{equation}
    \tau_{\theta}(y_w, y_l) = \min(e^{\eta\cdot z_{\theta}(y_w, y_l)} - 1 ,1),
\label{eq:tau}
\end{equation}
where $\eta$ is a hyperparameter that controls the impact of $z_{\theta}$, and $z_{\theta}(y_w, y_l)$ encodes the distance between the chosen and rejected responses by measuring the absolute difference in log-likelihoods:
\begin{equation} \label{eq:z_arpo}
    z_{\theta}(y_w, y_l)\!=\!\left|\frac{\log(\pi_{\theta}(y_w | x))}{|y_w|} - \frac{\log(\pi_{\theta}(y_l | x))}{|y_l|}\right|\!.    
\end{equation}
For curriculum learning, we employ $s\!=\!-z_{\theta}$ in  step 7 of Algorithm \ref{alg:clewr}. This version is called CLewR-$z$.

\noindent\textbf{Enhanced ARPO.}
We further introduce an enhanced variant of ARPO by using a different distance function $z'_{\theta}(y_w, y_l)$ that also accounts for distances in the evaluation metric spaces. Specifically, we use:
\begin{equation}
z'\!=\!\eta_1 \cdot z_{\theta} + \eta_2 \cdot z_{\text{BLEU}} + \eta_3 \cdot z_{\text{COMET}},
\end{equation}
where $z_\theta$ is the original distance. For both metrics, $z_{\text{metric}}$ is given by $1-\frac{\text{metric}}{100}$ to normalize each metric to a $(0,1)$ interval and have the same monotony as the original $z_\theta$. Two dissimilar predictions result in low BLEU and COMET values, so $z_{\text{BLEU}}$ and $z_{\text{COMET}}$ will be high. Each $z$ is multiplied by a corresponding scalar $\eta$, scaling them to similar intervals. We create multiple versions of enhanced ARPO by modifying the scalars $\eta_1$, $\eta_2$ and $\eta_3$. All ARPO versions are listed in Table \ref{tab:arpo_hparams}.

\vspace{-0.1cm}
\section{Experimental Setup}
\vspace{-0.1cm}

\noindent\textbf{Dataset.} We test on the Flores-200 \citep{costa2022no} dataset. For generic LLMs, we use a group of six Romance languages. For MT-adapted models (GemmaX2), we use a group of three Romance languages, following \citet{cui2025multilingual}. We select Chinese to showcase generalization beyond Romance languages.

\noindent\textbf{LLM backbones.} % what are the baselines PO / LLMs considered.
We consider several candidate LLMs for preference tuning: LLama3.1-8B \citep{grattafiori2024llama}, Qwen2.5-7B \citep{qwen2025qwen25technicalreport}, Gemma2-9B \citep{team2024gemma}, and GemmaX2-9B \citep{cui2025multilingual}. % which is based on Gemma2-9B. 
We also consider X-ALMA \citep{xu2024x} as a reference baseline, which is based on Llama2 \citep{touvron2023llama}. %, with separate low-rank adapters \citep{Hu-ICLR-2022} for different language groups.

\noindent\textbf{Preference optimization baselines.} % what are the baselines PO / LLMs considered.
We train the models using three different PO algorithms: Direct Preference Optimization Positive (DPOP) \citep{pal2024smaug} (an enhanced version of DPO \citep{rafailov2023direct}), CPO with behavior cloning \citep{xu2024contrastive}, and ARPO \citep{xu2024x}. We employ these algorithms with and without our CLewR. For ARPO, we integrate both CLewR and CLewR-$z$.

Another comparison is between CurriDPO~\citep{pattnaik2024enhancing} and CLewR. For CurriDPO, we use the iterative approach based on the same SFT reference model.

\begin{table}[!t]
\centering
\setlength\tabcolsep{0.25em}
\resizebox{1.0\linewidth}{!}{
\begin{tabular}{lcccc}
\toprule
\multirow{3.5}{*}{Model} 
& \multicolumn{4}{c}{\textbf{Test on 6 languages (Romance group)}} 
\\ \cmidrule{2-5}
& \multicolumn{2}{c}{\texttt{en}$\rightarrow$\texttt{xx}} 
& \multicolumn{2}{c}{\texttt{xx}$\rightarrow$\texttt{en}} 
\\
& BLEU $\uparrow$ & COMET $\uparrow$ & BLEU $\uparrow$ & COMET $\uparrow$
\\ \midrule

X-ALMA (ARPO)
& \textbf{39.83} & \textbf{89.38} & \textbf{42.88} & \textbf{89.28}
\\ \midrule

Gemma2-9B
& 19.85 & 61.92 & 34.98 & 85.57 
\\ \midrule

\quad + DPOP               
& 23.26$_{\pm 4.23}$ & 85.94$_{\pm 0.59}$ & 33.17$_{\pm 1.47}$ & 87.77$_{\pm 0.02}$ 
\\

\quad\quad+ CurriDPO 
& 21.81$_{\pm 0.26}$ & 85.64$_{\pm 0.34}$ & 28.33$_{\pm 2.71}$  & 87.27$_{\pm 0.20}$ 
\\

\quad\quad+ CLewR           
& 22.35$_{\pm 1.85}$ & 86.11$_{\pm 0.18}$ & 31.86$_{\pm 1.40}$  & 87.65$_{\pm 0.17}$ 
\\ \hdashline

\quad + CPO             
& 33.53$_{\pm 0.09}$ & 87.43$_{\pm 0.05}$ & 36.44$_{\pm 0.06}$ & 87.76$_{\pm 0.03}$ 
\\

\quad\quad+ CLewR$^\dagger$       
& 36.24$_{\pm 0.03}$ & 87.85$_{\pm 0.01}$ & 40.33$_{\pm 0.23}$ & 88.49$_{\pm 0.04}$ 
\\ \hdashline

\quad + ARPO               
& 35.37$_{\pm 0.02}$ & 88.12$_{\pm 0.04}$ & 37.68$_{\pm 0.63}$ & 88.30$_{\pm 0.10}$
\\

\quad\quad+ CLewR$^\dagger$             
& 36.63$_{\pm 0.05}$ & 88.12$_{\pm 0.01}$ & 40.44$_{\pm 0.06}$ & 88.56$_{\pm 0.02}$ 
\\

\quad\quad+ CLewR-$z^\dagger$            
& 36.50$_{\pm 0.02}$ & 88.12$_{\pm 0.03}$ & 39.87$_{\pm 0.34}$ & 88.62$_{\pm 0.06}$ 
\\ \hdashline

\quad + ARPO-$z'$-V1            
& 36.35$_{\pm 0.01}$ & 88.40$_{\pm 0.02}$ & 40.32$_{\pm 0.09}$ & 88.62$_{\pm 0.03}$ 
\\

\quad\quad+ CLewR$^\dagger$          
& \textbf{37.45}$_{\pm 0.02}$ & 88.40$_{\pm 0.04}$ & 41.28$_{\pm 0.11}$ & 88.78$_{\pm 0.03}$ 
\\

\quad\quad+ CLewR-$z^\dagger$            
& 36.98$_{\pm 0.03}$ & 88.38$_{\pm 0.04}$ & 41.22$_{\pm 0.08}$ & 88.84$_{\pm 0.02}$
\\ \hdashline

\quad + ARPO-$z'$-V2           
& 36.46$_{\pm 0.05}$ & \textbf{88.43}$_{\pm 0.03}$ & 40.10$_{\pm 0.16}$ & 88.58$_{\pm 0.05}$
\\

\quad\quad+ CLewR$^\dagger$            
& \textbf{37.45}$_{\pm 0.04}$ & 88.37$_{\pm 0.05}$ & 41.09$_{\pm 0.21}$ & 88.75$_{\pm 0.05}$ 
\\

\quad\quad+ CLewR-$z^\dagger$ 
& 37.08$_{\pm 0.02}$ & 88.38$_{\pm 0.01}$ & \textbf{41.45}$_{\pm 0.08}$ & \textbf{88.87}$_{\pm 0.02}$
\\ \midrule

Qwen2.5-7B              
& 4.10 & 45.94 & 9.59 & 62.83 
\\ \midrule

\quad + DPOP               
& 24.43$_{\pm 0.88}$ & 82.90$_{\pm 0.24}$ & 33.06$_{\pm 0.76}$ & 87.42$_{\pm 0.12}$
\\

\quad\quad+ CurriDPO 
& 23.80$_{\pm 0.26}$ & 81.90$_{\pm 0.30}$ & 29.12$_{\pm 2.58}$ & 86.77$_{\pm 0.39}$ 
\\ 

\quad\quad+ CLewR           
& 23.59$_{\pm 1.63}$ & 82.74$_{\pm 0.41}$ & 30.63$_{\pm 1.97}$ & 87.10$_{\pm 0.27}$
\\ \hdashline

\quad + CPO             
& 27.68$_{\pm 0.03}$ & 83.93$_{\pm 0.06}$ & 31.66$_{\pm 0.03}$ & 86.36$_{\pm 0.03}$
\\

\quad\quad+ CLewR$^\dagger$            
& 30.05$_{\pm 0.02}$ & 84.15$_{\pm 0.06}$ & 36.28$_{\pm 0.10}$ & 87.33$_{\pm 0.05}$
\\ \hdashline

\quad + ARPO               
& 30.41$_{\pm 0.05}$ & 85.64$_{\pm 0.04}$ & 32.12$_{\pm 0.10}$ & 86.71$_{\pm 0.04}$
\\

\quad\quad+ CLewR$^\dagger$            
& 31.56$_{\pm 0.10}$ & 85.38$_{\pm 0.07}$ & 36.69$_{\pm 0.27}$ & 87.62$_{\pm 0.10}$
\\

\quad\quad+ CLewR-$z^\dagger$             
& 31.14$_{\pm 0.06}$ & 85.47$_{\pm 0.04}$ & 34.88$_{\pm 0.08}$ & 87.45$_{\pm 0.02}$
\\ \hdashline

\quad + ARPO-$z'$-V1           
& 31.08$_{\pm 0.14}$ & 85.67$_{\pm 0.01}$ & 35.12$_{\pm 0.91}$ & 87.37$_{\pm 0.29}$
\\

\quad\quad+ CLewR$^\dagger$            
& 32.68$_{\pm 0.02}$ & 85.81$_{\pm 0.01}$ & \textbf{37.18}$_{\pm 0.08}$ & 87.60$_{\pm 0.03}$
\\

\quad\quad+ CLewR-$z^\dagger$             
& 32.68$_{\pm 0.05}$ & \textbf{85.93}$_{\pm 0.05}$ & 36.55$_{\pm 0.01}$ & 87.56$_{\pm 0.02}$
\\ \hdashline

\quad + ARPO-$z'$-V2          
& 30.95$_{\pm 0.09}$ & 85.67$_{\pm 0.03}$ & 36.40$_{\pm 0.10}$ & \textbf{87.75}$_{\pm 0.02}$
\\

\quad\quad+ CLewR$^\dagger$            
& \textbf{32.70}$_{\pm 0.03}$ & 85.78$_{\pm 0.01}$ & 37.16$_{\pm 0.06}$ & 87.64$_{\pm 0.03}$
\\

\quad\quad+ CLewR-$z$            
& 32.67$_{\pm 0.04}$ & \textbf{85.93}$_{\pm 0.02}$ & 36.15$_{\pm 0.08}$ & 87.51$_{\pm 0.02}$
\\ \midrule

LLaMA3.1-8B              
& 1.70 & 46.00 & 1.84 & 46.30 
\\ \midrule

\quad + DPOP               
& 31.53$_{\pm 0.11}$ & 87.05$_{\pm 0.08}$ & 35.78$_{\pm 0.46}$ & 87.82$_{\pm 0.01}$
\\

\quad\quad+ CurriDPO 
& 32.46$_{\pm 0.36}$ & 86.94$_{\pm 0.06}$ & 36.01$_{\pm 0.50}$ & 87.75$_{\pm 0.05}$ 
\\ 

\quad\quad+ CLewR           
& 30.87$_{\pm 0.37}$ & 86.85$_{\pm 0.07}$ & 35.46$_{\pm 0.65}$ & 87.83$_{\pm 0.03}$
\\ \hdashline

\quad + CPO             
& 1.47$_{\pm 0.34}$ & 54.97$_{\pm 2.52}$ & 10.36$_{\pm 4.48}$ & 74.32$_{\pm 1.78}$
\\

\quad\quad+ CLewR$^\dagger$            
& 3.97$_{\pm 0.99}$ & 59.66$_{\pm 3.21}$ & 13.08$_{\pm 1.93}$ & 76.76$_{\pm 3.90}$
\\ \hdashline

\quad + ARPO               
& 33.50$_{\pm 0.20}$ & 87.48$_{\pm 0.05}$ & 35.92$_{\pm 0.56}$ & 87.75$_{\pm 0.11}$
\\

\quad\quad+ CLewR$^\dagger$            
& 34.49$_{\pm 0.05}$ & 87.44$_{\pm 0.06}$ & 37.99$_{\pm 0.07}$ & 87.94$_{\pm 0.04}$
\\

\quad\quad+ CLewR-$z^\dagger$              
& 34.33$_{\pm 0.09}$ & 87.57$_{\pm 0.04}$ & 36.75$_{\pm 0.17}$ & 87.93$_{\pm 0.03}$
\\ \hdashline

\quad + ARPO-$z'$-V1           
& 33.68$_{\pm 0.15}$ & 87.44$_{\pm 0.02}$ & 37.82$_{\pm 0.35}$ & 87.91$_{\pm 0.05}$
\\

\quad\quad+ CLewR$^\dagger$            
& 35.49$_{\pm 0.04}$ & 87.77$_{\pm 0.01}$ & \textbf{39.05}$_{\pm 0.07}$ & \textbf{88.22}$_{\pm 0.01}$
\\

\quad\quad+ CLewR-$z^\dagger$               
& 35.34$_{\pm 0.05}$ & 87.77$_{\pm 0.03}$ & 38.02$_{\pm 0.11}$ & 88.15$_{\pm 0.02}$
\\ \hdashline

\quad + ARPO-$z'$-V2          
& 33.96$_{\pm 0.09}$ & 87.49$_{\pm 0.10}$ & 37.47$_{\pm 0.15}$ & 87.84$_{\pm 0.06}$
\\

\quad\quad+ CLewR$^\dagger$            
& \textbf{35.53}$_{\pm 0.03}$ & 87.77$_{\pm 0.04}$ & 38.89$_{\pm 0.07}$ & 88.17$_{\pm 0.03}$
\\

\quad\quad+ CLewR-$z^\dagger$ 
& 35.43$_{\pm 0.01}$ & \textbf{87.80}$_{\pm 0.04}$ & 37.87$_{\pm 0.09}$ & 88.13$_{\pm 0.04}$
\\ 
\bottomrule
\end{tabular}
}
\vspace{-0.2cm}
\caption{
Results across several base models and preference optimization methods, with and without curriculum learning. Best overall results and per-model best results are shown in bold. A dagger ($\dagger$) indicates statistical significance over the non-curriculum PO method.}
\vspace{-0.25cm}
\label{tab:group2}
\end{table}

% the data is split into three levels of difficulty (easy, medium and hard). % and trained sequentially first with the shuffled easy data and then with the medium and hard, 3 epochs each. 

\noindent\textbf{Hyperparameter tuning.} % batch size, learning rate, LoRA params, optimizer, etc.
We fine-tune LoRA \cite{Hu-ICLR-2022} adapters on top of the base models, using a rank of $64$ for the LoRA matrices, a learning rate of $5\cdot10^{-5}$, a warmup ratio of $0.1$, and a mini-batch size of $4$. All models are optimized with Adam \cite{Kingma-ICLR-1015}. Gemma2, Qwen2.5 and Llama3.1 are each trained for three epochs. Since GemmaX2 is already tuned for MT, it is trained in one epoch. We use default values for other hyperparameters.

% For the original $z$ in ARPO, we use $\eta=1.5$. For $z_{\text{BLEU}}$ and $z_{\text{COMET}}$, we use $\eta_{\text{BLEU}}=0.1$ and $\eta_{\text{COMET}}=0.5$.

\noindent\textbf{Training and evaluation protocol.} % what metrics, how experiments are organized
We focus on translation to and from English for languages from the Romance language group used in X-ALMA \citep{xu2024x}: Catalan (\texttt{ca}), Galician (\texttt{gl}), Italian (\texttt{it}), Portuguese (\texttt{pt}), Romanian (\texttt{ro}) and Spanish (\texttt{es}). We use the X-ALMA\footnote{\href{https://huggingface.co/datasets/haoranxu/X-ALMA-Preference}{XALMA} preference dataset.} preference data (81K train and 7K validation) and select the triplets from the \texttt{en}$\rightarrow$\texttt{xx} and \texttt{xx}$\rightarrow$\texttt{en} language pairs, where \texttt{xx} belongs to the Romance language group. For Gemma2-9B, Qwen2.5-7B and LLama3.1-8B, we use all six languages, while for GemmaX2, we select $\{\texttt{it}, \texttt{pt}, \texttt{es}\}$, since the model was not adapted for MT on the other three languages. %and therefore a further preference optimization on all of them would not be appropriate for achieving the optimal results.

% add here intro about chinese?
Furthermore, to assess the generalization of our method beyond Romance languages, we also report results on a linguistically distant language pair. Specifically, we evaluate Gemma2-9B on Chinese (\texttt{zh}) using the X-ALMA preference dataset, considering both \texttt{en}$\rightarrow$\texttt{zh} and \texttt{zh}$\rightarrow$\texttt{en} directions.

To evaluate the models, we compute BLEU and COMET-22 across all considered languages for both \texttt{en}$\rightarrow$\texttt{xx} and \texttt{xx}$\rightarrow$\texttt{en} directions. For each setup, we perform three training runs and report the average scores and the corresponding standard deviations across runs. For each run, we select the checkpoint with the best validation performance.

To complement BLEU and COMET, we report additional metrics in the results. Specifically, we use the corpus-level Moving Average Type-Token Ratio (MATTR)~\cite{covington2010cutting} to measure lexical diversity, as well as TER~\cite{DBLP:conf/amta/SnoverDSMM06}, chrF~\cite{DBLP:conf/wmt/Popovic15}, and an LLM-as-a-judge metric~\cite{DBLP:conf/nips/ZhengC00WZL0LXZ23} to assess translation quality. TER measures edit distance, while chrF captures character n-gram overlap, being more robust to morphological variation. The LLM-as-a-judge prompt is given in Appendix \ref{sec:llm-as-a-judge}.

\vspace{-0.1cm}
\section{Results}
\vspace{-0.1cm}

We focus on two main comparisons: CLewR vs.~CurriDPO, and CLewR vs.~non-curriculum. In Table~\ref{tab:group2}, we present MT results with various PO methods, with and without curriculum, applied on generic LLMs (Gemma2, Qwen2.5 and Llama3.1). 

\noindent\textbf{CLewR vs.~CurriDPO.}
Across all generic LLMs, DPOP+CLewR outperforms DPOP+CurriDPO in terms of COMET. The improvements, as measured via paired bootstrap resampling with 10,000 samples and a $95\%$ confidence interval, are statistically significant on 2 out of the 3 models (Gemma2 and Qwen2.5).  However, CLewR does not boost the performance of the standalone DPOP. We hypothesize that this is due to the DPOP margin, which depends on both policy $\pi_{\theta}$ and reference model $\pi_{\theta_{\text{ref}}}$:  $m=m(\pi_{\theta})-m(\pi_{\theta_{\text{ref}}})$. Because the reference margin is typically positive, it offsets part of the policy margin. As a result, the optimization signal in DPOP is weaker compared to other preference optimization methods that rely only on the policy margin. Indeed, CLewR brings consistent performance gains to highly competitive preference optimization methods, namely CPO and ARPO, which rely on a different \emph{modus operandi}.

\noindent\textbf{CLewR vs.~base PO methods.}
CLewR consistently improves both BLEU and COMET scores across LLMs and PO algorithms. Most improvements are statistically significant according to paired bootstrap resampling with a $95\%$ confidence interval over all sentence pairs. % Although gains are statistically significant in nearly all configurations, effect sizes are generally small, reflecting the incremental nature of improvements in large-scale MT.
Moreover, our ARPO-$z'$-V1 and ARPO-$z'$-V2 variants further improve performance over the other preference methods. This indicates that by incorporating external semantic signals derived from MT metrics into $z'$, our versions of ARPO better adjust preference margins. %according to semantic differences between response pairs, resulting in better-aligned and more stable optimization.  
ARPO-$z'$-V2 obtains the best performance on average. Additional experiments exploring more $z'$ combinations are presented in Appendix \ref{sec:ablation}, for \texttt{en}$\rightarrow$\texttt{ro} translations.

\noindent\textbf{MT-adapted LLM.} In Table~\ref{tab:gemmax2}, we compare the current state-of-the-art LLM for MT, GemmaX2, with PO-based variants and observe that the current SOTA can be further improved by applying preference optimization or curriculum learning. However, these gains are less pronounced than those previously observed in Table~\ref{tab:group2}, and the strongest overall performance for GemmaX2 is obtained using DPOP. This suggests that LLMs optimized for MT may benefit less from more complex preference optimization methods and curriculum learning.

%Statistical significance was assessed using paired bootstrap resampling, from which $95\%$ confidence intervals were computed.
%When averaging BLEU and COMET scores across all configurations, both directions, CLewR trainings yield a statistically significant improvement over the corresponding non-curriculum variants. 

\begin{table}[!t]
%\vskip 0.05in
\centering
\setlength\tabcolsep{0.2em}
\resizebox{1.0\linewidth}{!}{
\begin{tabular}{lcccc}
\toprule
\multirow{3.5}{*}{Model} 
& \multicolumn{4}{c}{\textbf{Test on 3 languages (\texttt{es}, \texttt{it}, \texttt{pt})}} 
\\ \cmidrule{2-5}
& \multicolumn{2}{c}{\texttt{en}$\rightarrow$\texttt{xx}} 
& \multicolumn{2}{c}{\texttt{xx}$\rightarrow$\texttt{en}} 

\\
& BLEU $\uparrow$ & COMET $\uparrow$ & BLEU $\uparrow$ & COMET $\uparrow$
\\ \midrule

X-ALMA (ARPO)
& 37.50 & 89.00 & 39.40 & 88.80
\\ \midrule

Gemma-X2-9B               
& 37.27 & 88.85 & 41.06 & 88.87 
\\ \midrule

\quad + DPOP               
& 38.23$_{\pm 0.01}$ & 89.01$_{\pm 0.02}$ & \textbf{41.93}$_{\pm 0.10}$ & 89.09$_{\pm 0.02}$
\\

\quad\quad+ CurriDPO
& 38.01$_{\pm 0.01}$ & 89.04$_{\pm 0.02}$ & 41.60$_{\pm 0.05}$ & 89.08$_{\pm 0.00}$ 
\\ 

\quad\quad+ CLewR             
& 38.24$_{\pm 0.16}$ & 88.93$_{\pm 0.01}$ & 41.80$_{\pm 0.07}$ & 89.09$_{\pm 0.02}$
\\ \hdashline

\quad + CPO             
& 36.36$_{\pm 0.04}$ & 89.04$_{\pm 0.00}$ & 41.00$_{\pm 0.09}$ & 89.05$_{\pm 0.01}$
\\

\quad\quad+ CLewR$^\dagger$            
& 37.69$_{\pm 0.03}$ & 88.87$_{\pm 0.01}$ & 41.45$_{\pm 0.14}$ & 89.04$_{\pm 0.01}$
\\ \hdashline

\quad + ARPO               
& 38.50$_{\pm 0.03}$ & \textbf{89.14}$_{\pm 0.01}$ & 40.45$_{\pm 0.02}$ & 89.06$_{\pm 0.00}$
\\

\quad\quad+ CLewR$^\dagger$            
& 37.83$_{\pm 0.01}$ & 88.93$_{\pm 0.01}$ & 41.46$_{\pm 0.02}$ & 89.06$_{\pm 0.00}$
\\

\quad\quad+ CLewR-$z$            
& 38.35$_{\pm 0.01}$ & 89.12$_{\pm 0.00}$ & 40.29$_{\pm 0.02}$ & 89.02$_{\pm 0.01}$

\\ \hdashline

\quad + ARPO-$z'$-V1           
& 38.56$_{\pm 0.04}$ & 89.07$_{\pm 0.01}$ & 41.31$_{\pm 0.13}$ & 89.04$_{\pm 0.02}$
\\

\quad\quad+ CLewR            
& 38.08$_{\pm 0.01}$ & 88.98$_{\pm 0.00}$ & 41.72$_{\pm 0.02}$ & 89.09$_{\pm 0.00}$
\\

\quad\quad+ CLewR-$z$           
& 38.26$_{\pm 0.01}$ & 89.09$_{\pm 0.01}$ & 40.87$_{\pm 0.22}$ & 89.02$_{\pm 0.01}$
\\ \hdashline

\quad + ARPO-$z'$-V2           
& \textbf{38.58}$_{\pm 0.04}$ & 89.07$_{\pm 0.00}$ & 41.33$_{\pm 0.02}$ & 89.04$_{\pm 0.00}$
\\

\quad\quad+ CLewR            
& 38.06$_{\pm 0.02}$ & 88.99$_{\pm 0.01}$ & 41.83$_{\pm 0.01}$ & \textbf{89.10}$_{\pm 0.01}$
\\

\quad\quad+ CLewR-$z$            
& 38.24$_{\pm 0.03}$ & 89.08$_{\pm 0.01}$ & 40.91$_{\pm 0.25}$ & 89.02$_{\pm 0.03}$

\\ \bottomrule

\end{tabular}
}
\vspace{-0.2cm}
\caption{Results of GemmaX2 combined with preference optimization methods, with and without curriculum. The new variations of GemmaX2 surpass both the original model GemmaX2 and X-ALMA. The best results for BLEU and COMET across \texttt{en}$\rightarrow$\texttt{xx} and \texttt{xx}$\rightarrow$\texttt{en} directions are in bold. A dagger ($\dagger$) indicates statistical significance over the non-curriculum PO method. }
\vspace{-0.25cm}
\label{tab:gemmax2}
\end{table}

%observations:s
% - preference optimization SOTA
% - curriculum helps preference optimizastion but is not SOTA

% chinese 
\noindent\textbf{Chinese results.} In Table~\ref{tab:a3}, we report further experiments for \texttt{en}$\rightarrow$\texttt{zh} and \texttt{zh}$\rightarrow$\texttt{en} language pairs. Our curriculum strategy always improves both ARPO-$z'$ variants on both directions and metrics.

\begin{table}[!t]
\centering
\setlength\tabcolsep{0.2em}
\resizebox{1.0\linewidth}{!}{
\begin{tabular}{lcccc}
\toprule
\multirow{3.5}{*}{Model} 
& \multicolumn{4}{c}{\textbf{Test on English$\leftrightarrow$Chinese}} 
\\ \cmidrule{2-5}
& \multicolumn{2}{c}{\texttt{zh}$\rightarrow$\texttt{en}} 
& \multicolumn{2}{c}{\texttt{en}$\rightarrow$\texttt{zh}} 
\\
& BLEU $\uparrow$ & COMET $\uparrow$ & BLEU $\uparrow$ & COMET $\uparrow$ \\
\midrule 
Gemma2-9B            & 27.84 & 86.90 & 38.27 & 81.18 \\
\midrule
\quad + ARPO-$z'$-V1      & 27.66$_{\pm 0.07}$ & 87.15$_{\pm 0.07}$ & 41.34$_{\pm 0.26}$ & 88.15$_{\pm 0.06}$ \\
\quad\quad+ CLewR-$z$  & \textbf{29.49$_{\pm 0.02}$} & \textbf{87.43$_{\pm 0.04}$} & \textbf{42.46$_{\pm 0.10}$} & \textbf{88.22$_{\pm 0.02}$} \\
\hdashline
\quad + ARPO-$z'$-V2      & 27.94$_{\pm 0.26}$ & 87.20$_{\pm 0.03}$ & 41.14$_{\pm 0.13}$ & 88.12$_{\pm 0.06}$ \\
\quad\quad+ CLewR-$z$  & \textbf{29.15$_{\pm 0.35}$} & \textbf{87.40$_{\pm 0.06}$} & \textbf{42.75$_{\pm 0.14}$} & \textbf{88.20$_{\pm 0.06}$} \\
\bottomrule
\end{tabular}}
\vspace{-0.2cm}
\caption{English$\leftrightarrow$Chinese results of Gemma2-9B based on ARPO-$z'$, with and without curriculum. The results are averaged over three runs.}
\label{tab:a3}
\vspace{-0.25cm}
\end{table}

\begin{table}[t]
\centering
\footnotesize
\setlength{\tabcolsep}{2.5pt}
\resizebox{1.0\linewidth}{!}{
\begin{tabular}{lcccccc}
\toprule
\multirow{2.5}{*}{Model} & \multicolumn{2}{c}{MATTR $\uparrow$} & \multicolumn{2}{c}{TER $\downarrow$} & \multicolumn{2}{c}{chrF $\uparrow$} \\
\cmidrule(lr){2-3} \cmidrule(lr){4-5} \cmidrule(lr){6-7}
 & \texttt{en}$\rightarrow$\texttt{xx} & \texttt{xx}$\rightarrow$\texttt{en} & \texttt{en}$\rightarrow$\texttt{xx} & \texttt{xx}$\rightarrow$\texttt{en} & \texttt{en}$\rightarrow$\texttt{xx} & \texttt{xx}$\rightarrow$\texttt{en} \\
\midrule

Gemma2-9B        & 85.87 & 83.15 & 91.29 & 61.31 & 39.11 & 65.66 \\
\midrule

\quad+ DPO        & \textbf{69.10} & \textbf{78.97} & \textbf{102.90} & \textbf{61.82} & \textbf{54.48} & \textbf{62.62} \\
\quad\quad+ curriDPO & 67.70 & 71.92 & 107.03 & 82.23 & 54.44 & 60.56 \\
\quad\quad+ CLewR    & 68.62 & 76.96 & 103.15 & 68.61 & 54.38 & 61.98 \\
\hdashline
\quad+ CPO          & \textbf{83.15} & \textbf{82.64} & 54.36 & 51.20 & 60.07 & 63.04 \\
\quad\quad+ CLewR    & 83.14 & 82.60 & \textbf{50.00} & \textbf{46.09} & \textbf{62.21} & \textbf{65.61} \\
\hdashline
\quad+ ARPO         & 83.06 & \textbf{82.75} & 51.02 & 49.18 & 61.49 & 64.16 \\
\quad\quad+ CLewR    & 83.33 & 82.65 & \textbf{49.43} & \textbf{45.88} & \textbf{62.50} & 65.73 \\
\quad\quad+ CLewR-$z$  & \textbf{83.53} & \textbf{82.75} & 49.81 & 46.74 & 62.41 & \textbf{65.74} \\
\hdashline
\quad+ ARPO-$z'$-V1   & 82.41 & 82.34 & 50.59 & 46.37 & 62.25 & 65.71 \\
\quad\quad+ CLewR    & \textbf{83.18} & \textbf{82.79} & \textbf{48.70} & \textbf{45.03} & \textbf{63.00} & 66.43 \\
\quad\quad+ CLewR-$z$  & 82.68 & 82.46 & 49.52 & 45.38 & 62.67 & \textbf{66.47} \\
\hdashline
\quad+ ARPO-$z'$-V2   & 82.56 & 82.41 & 50.34 & 46.78 & 62.37 & 65.54 \\
\quad\quad+ CLewR    & \textbf{83.18} & \textbf{82.82} & \textbf{48.67} & \textbf{45.21} & \textbf{63.05} & 66.42 \\
\quad\quad+ CLewR-$z$  & 82.77 & 82.38 & 49.34 & 45.18 & 62.74 & \textbf{66.61} \\

\bottomrule
\end{tabular}
}
\vspace{-0.2cm}
\caption{Diversity (MATTR) and performance (TER, chrF) metrics for Gemma2-9B. For each PO method, the best score is in bold.}
\label{tab:a1}
\vspace{-0.25cm}
\end{table}

\noindent\textbf{Diversity metrics.}
We further report additional results in terms of diversity and translation quality metrics in Tables~\ref{tab:a1} and~\ref{tab:llm_judge}, respectively. Consistent with the trends observed for BLEU and COMET, CLewR generally provides improvements for the more advanced preference optimization methods in terms of lexical diversity (MATTR), translation quality (chrF), while also reducing edit distance as measured by TER. This indicates that the generated translations require fewer post-edits to match the reference, reflecting better overall adequacy and fluency alignment. We additionally analyze Borda count rankings in Appendix \ref{sec:borda}, where we provide a more detailed breakdown across individual language pairs and translation directions.
%The degradation in TER score may be explained by the fact that the increased output diversity leads to more word reorderings, which are penalized by edit distance based methods. }

\noindent\textbf{LLM-as-a-Judge.} We report LLM-as-a-judge evaluations in Appendix \ref{sec:llm-as-a-judge}.

\noindent\textbf{Ablation study.} We discuss ablation experiments in Appendix \ref{sec:ablation}.

\vspace{-0.1cm}
\section{Conclusion}
\vspace{-0.1cm}

In this paper, we introduced a novel curriculum learning strategy for preference optimization in MT. We demonstrated consistent gains across several general LLM families. We also showed that the adaptive penalty in ARPO based on modified distances $z'$ derived from MT metrics yields additional performance gains. Moreover, we found that CLewR-$z$ can further improve performance over the original curriculum scoring function. 

In future work, we aim to explore the application of CLewR to other NLP tasks.

\vspace{-0.1cm}
\section*{Acknowledgments}
\vspace{-0.1cm}

This work was supported by a grant of the Ministry of Research, Innovation and Digitization, CNCS -
UEFISCDI, project number PN-IV-P1-PCE-2023-0354, within PNCDI IV.

\section*{Limitations}
\vspace{-0.1cm}

Due to computational constraints, we limited our experiments to the 6 languages in the Romance language group as defined in X-ALMA. Generalization is demonstrated only on Chinese. More experiments on other language groups and other model families would help strengthen the results.

Similar to generic PO algorithms, CLewR does not specifically address training data and model-specific biases. To mitigate potential biases, additional debiasing techniques need to be applied.

\bibliography{custom}

\appendix

\begin{table}[!t]
\centering
\resizebox{0.92\linewidth}{!}{
\begin{tabular}{lcc}
\toprule
\multirow{3.5}{*}{Model} 
& \multicolumn{2}{c}{\textbf{Test on Romanian}} 
\\ \cmidrule{2-3}
& \multicolumn{2}{c}{\texttt{en}$\rightarrow$\texttt{ro}} 

\\
& BLEU $\uparrow$ & COMET $\uparrow$
\\ \midrule

Gemma2-9B
& 23.25 & 68.59
\\ \midrule

\phantom{  +} + ARPO               
& 38.18$_{\pm 0.55}$ & 90.54$_{\pm 0.02}$
\\

\phantom{pholder}+ CLewR            
& 38.53$_{\pm 0.30}$ & 90.58$_{\pm 0.01}$
\\

\phantom{pholder}+ CLewR-$z$            
& 38.49$_{\pm 0.20}$ & 90.49$_{\pm 0.05}$
\\ \hdashline

\phantom{  +} + ARPO $\eta=0.1$           
& 38.29$_{\pm 0.03}$ & 90.63$_{\pm 0.01}$
\\

\phantom{pholder}+ CLewR            
& \textbf{39.78}$_{\pm 0.02}$ & 90.79$_{\pm 0.01}$
\\

\phantom{pholder}+ CLewR-$z$           
& 39.10$_{\pm 0.03}$ & 90.61$_{\pm 0.01}$
\\ \hdashline

\phantom{  +} + ARPO $\beta=0.5$           
& 35.62$_{\pm 0.10}$ & 90.26$_{\pm 0.03}$
\\

\phantom{pholder}+ CLewR            
& 37.31$_{\pm 0.01}$ & 90.08$_{\pm 0.01}$
\\

\phantom{pholder}+ CLewR-$z$            
& 36.96$_{\pm 0.07}$ & 90.41$_{\pm 0.01}$
\\ \hdashline

% \phantom{  +} + ARPO log           
% & 35.10$_{\pm 0.25}$ & 90.12$_{\pm 0.23}$
% \\

% \phantom{pholder}+ curriculum            
% & 37.31$_{\pm 0.08}$ & 90.05$_{\pm 0.08}$
% \\

% \phantom{pholder}+ curriculum z'            
% & 36.88$_{\pm 0.20}$ & 90.35$_{\pm 0.01}$
% \\ \hdashline

% \phantom{  +} + ARPO log $\eta=0.1$           
% & 38.05$_{\pm 0.12}$ & 90.43$_{\pm 0.01}$
% \\

% \phantom{pholder}+ curriculum            
% & 38.67$_{\pm 0.05}$ & 90.69$_{\pm 0.00}$
% \\

% \phantom{pholder}+ curriculum z'            
% & 38.65$_{\pm 0.07}$ & 90.53$_{\pm 0.01}$
% \\ \hdashline

% \phantom{  +} + ARPO log reversed (10ep)           
% & 35.07 & 89.73
% \\

% \phantom{pholder}+ curriculum            
% & 35.97 & 90.24
% \\

% \phantom{pholder}+ curriculum z'            
% & 36.53 & 89.84
% \\ \hdashline

\phantom{  +} + ARPO-$z'$-V3            
& 36.98$_{\pm 0.22}$ & 90.23$_{\pm 0.02}$
\\

\phantom{pholder}+ CLewR            
& 38.43$_{\pm 0.01}$ & 90.32$_{\pm 0.01}$
\\

\phantom{pholder}+ CLewR-$z$        
& 38.58$_{\pm 0.01}$ & 90.53$_{\pm 0.01}$
\\ \hdashline

\phantom{  +} + ARPO-$z'$-V4            
& 37.29$_{\pm 0.14}$ & 90.34$_{\pm 0.05}$
\\

\phantom{pholder}+ CLewR            
& 38.44$_{\pm 0.05}$ & 90.46$_{\pm 0.02}$
\\

\phantom{pholder}+ CLewR-$z$           
& 38.46$_{\pm 0.04}$ & 90.39$_{\pm 0.01}$
\\ \hdashline

\phantom{  +} + ARPO-$z'$-V1           
& 38.33$_{\pm 0.26}$ & 90.59$_{\pm 0.00}$
\\

\phantom{pholder}+ CLewR            
& \textbf{39.71}$_{\pm 0.06}$ & 90.75$_{\pm 0.02}$
\\

\phantom{pholder}+ CLewR-$z$          
& 39.38$_{\pm 0.02}$ & 90.73$_{\pm 0.04}$
\\ \hdashline

\phantom{  +} + ARPO ARPO-$z'$-V5     
& 37.31$_{\pm 0.00}$ & 90.18$_{\pm 0.01}$
\\

\phantom{pholder}+ CLewR            
& 38.99$_{\pm 0.02}$ & 90.54$_{\pm 0.02}$
\\

\phantom{pholder}+ CLewR-$z$            
& 38.49$_{\pm 0.07}$ & 90.43$_{\pm 0.02}$
\\ \hdashline

\phantom{  +} + ARPO-$z'$-V6             
& 37.18$_{\pm 0.03}$ & 90.18$_{\pm 0.00}$
\\

\phantom{pholder}+ CLewR            
& 38.64$_{\pm 0.03}$ & 90.52$_{\pm 0.01}$
\\

\phantom{pholder}+ CLewR-$z$            
& 38.53$_{\pm 0.01}$ & 90.46$_{\pm 0.02}$
\\ \hdashline

\phantom{  +} + ARPO-$z'$-V7             
& 38.06$_{\pm 0.03}$ & 90.58$_{\pm 0.01}$
\\

\phantom{pholder}+ CLewR            
& 39.52$_{\pm 0.04}$ & 90.63$_{\pm 0.01}$
\\

\phantom{pholder}+ CLewR-$z$            
& 38.70$_{\pm 0.01}$ & 90.77$_{\pm 0.01}$
\\ \hdashline

\phantom{  +} + ARPO-$z'$-V8           
& 38.23$_{\pm 0.01}$ & 90.51$_{\pm 0.02}$
\\

\phantom{pholder}+ CLewR            
& 39.59$_{\pm 0.07}$ & 90.76$_{\pm 0.05}$
\\

\phantom{pholder}+ CLewR-$z$           
& 39.62$_{\pm 0.04}$ & \textbf{90.86}$_{\pm 0.01}$
\\ \hdashline

% \phantom{  +} + ARPO $z'=\frac{1}{2}z+\frac{1}{3}z_B+\frac{1}{6}z_C$           
% & 36.27$_{\pm 0.08}$ & 90.28$_{\pm 0.01}$
% \\

% \phantom{pholder}+ curriculum            
% & 38.37$_{\pm 0.01}$ & 90.35$_{\pm 0.01}$
% \\

% \phantom{pholder}+ curriculum z'            
% & 37.74$_{\pm 0.03}$ & 90.21$_{\pm 0.02}$
% \\ \hdashline

\phantom{  +} + ARPO-$z'$-V9             
& 36.79$_{\pm 0.08}$ & 90.30$_{\pm 0.09}$
\\

\phantom{pholder}+ CLewR            
& 38.43$_{\pm 0.06}$ & 90.41$_{\pm 0.05}$
\\

\phantom{pholder}+ CLewR-$z$         
& 37.98$_{\pm 0.15}$ & 90.31$_{\pm 0.07}$
\\ \hdashline

\phantom{  +} + ARPO-$z'$-V2           
& 38.16$_{\pm 0.11}$ & 90.50$_{\pm 0.02}$
\\

\phantom{pholder}+ CLewR            
& 39.66$_{\pm 0.09}$ & 90.72$_{\pm 0.01}$
\\

\phantom{pholder}+ CLewR-$z$            
& 39.58$_{\pm 0.03}$ & \textbf{90.86}$_{\pm 0.01}$

\\ \bottomrule

\end{tabular}
}
\vspace{-0.1cm}
\caption{
Ablation study on translations from English to Romanian, experimenting with different hyperparameters and variations of $z$ for ARPO.
}
%\vskip 0.05in
\label{tab:gemma2_ablations_ro}
\vspace{-0.1cm}
\end{table}

\begin{table}[t]
\centering
\renewcommand{\arraystretch}{1.2}
\begin{tabular}{lccc}
\toprule
ARPO Version & $\eta_1$ & $\eta_2$ & $\eta_2$ \\
\midrule
ARPO & $1.5 $& $0$ & $0$ \\
ARPO-$z'$-V1 & $0$ & $0$ & $0.5$ \\
ARPO-$z'$-V2 & $\frac{0.1}{3} $ & $\frac{0.1}{3}$ & $\frac{0.5}{3}$ \\
ARPO-$z'$-V3 & $0$ & $1.5$ & $0$  \\
ARPO-$z'$-V4 & $0$ & $0$ & $6$  \\
% ARPO-$z'$-V5 & $0$ & $0$ & $0.5$  \\
ARPO-$z'$-V5 & $0$ & $\frac{1.5}{2}$ & $\frac{6}{2}$ \\
ARPO-$z'$-V6 & $\frac{1.5}{2} $ & $\frac{1.5}{2}$ & $0$ \\
ARPO-$z'$-V7 & $\frac{1.5}{2} $ & $0$ & $\frac{6}{2}$ \\
ARPO-$z'$-V8 & $\frac{0.1}{2} $ & $0$ & $\frac{0.5}{2}$ \\
ARPO-$z'$-V9 & $\frac{1.5}{3} $ & $\frac{1.5}{3}$ & $\frac{6}{3}$ \\
% ARPO-$z'$-V11 & $\frac{0.1}{3} $ & $\frac{0.1}{3}$ & $\frac{0.5}{3}$ \\
\bottomrule
\end{tabular}
\vspace{-0.2cm}
\caption{ARPO variants and their hyperparameter configurations.}
\label{tab:arpo_hparams}
\vspace{-0.2cm}
\end{table}

\section{Appendix}
\label{sec:appendix}
\begin{figure*}[!tb]
    \centering
    \includegraphics[width=1.0\textwidth]{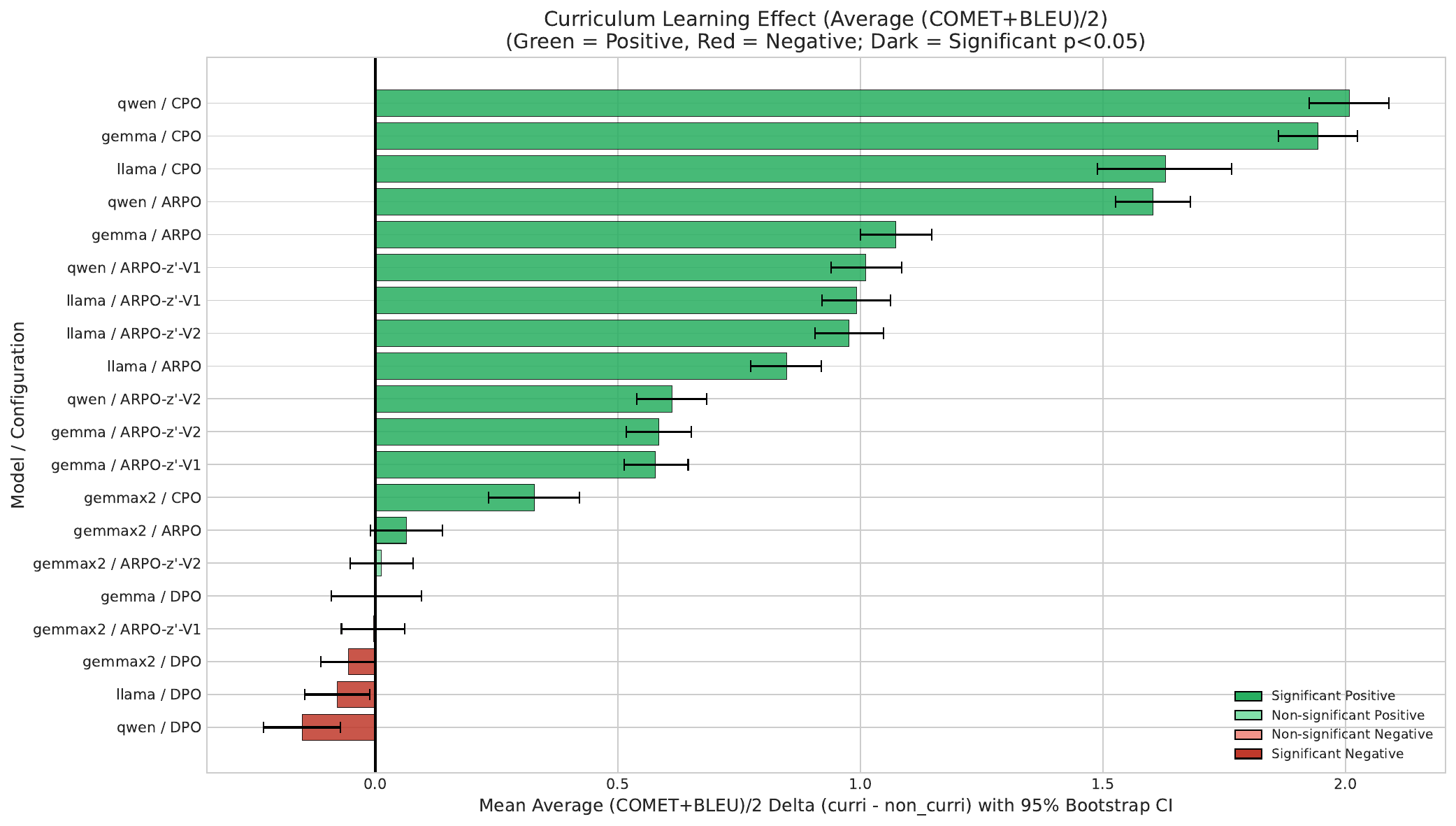}
    \vspace{-0.7cm}
    \caption{Tested configurations and their improvements from curriculum against the non-curriculum variant. 13 out of 20 configurations show statistically significant improvements, 4 show no significant change, and 3 result in performance degradation. The black bars represent $95\%$ confidence intervals. Best viewed in color.}
    \label{fig:configs_ranking}
    \vspace{-0.2cm}
\end{figure*}

\subsection{Ablation Study}
\label{sec:ablation}

In Table~\ref{tab:gemma2_ablations_ro}, we evaluate several $z'$ variants and hyperparameters ($\eta,\beta$). The hyperparameter configuration for every evaluated variant is given in Table \ref{tab:arpo_hparams}. The best variants of ARPO typically involve $z_{\text{COMET}}$. $z_{\text{BLEU}}$ helps performance when averaged with $z_\theta$ and $z_{\text{COMET}}$. Smaller values for $\eta_1$, $\eta_2$ and $\eta_3$ lead to better performance.

\subsection{Overview of Performance Gains}

In Figure \ref{fig:configs_ranking}, we present a detailed analysis on each configuration and its statistical improvements. For each model, the most significant increase is obtained by applying CLewR over CPO. CLewR over DPO degrades performance, suggesting that our curriculum is more beneficial for advanced preference algorithms.

\begin{figure*}[t]
\centering
\includegraphics[width=1.0\textwidth]{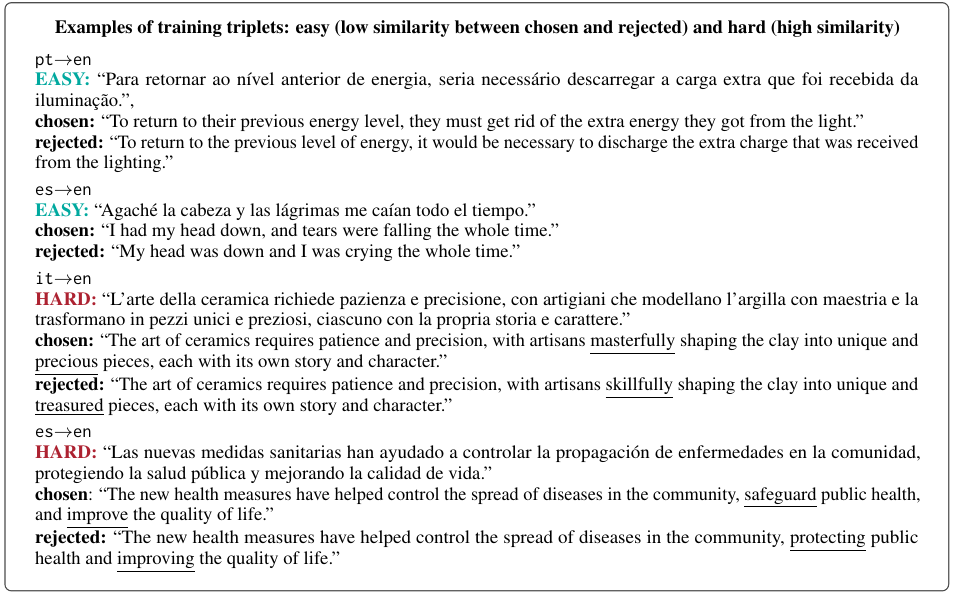}

\caption{Easy vs.~hard examples for curriculum learning based on the score $s$ derived from BLEU, METEOR and COMET metrics.}
\label{fig:examples}
\end{figure*}

% \subsection{Hyperparameters for variants of ARPO}
% \label{sec:examples}

\subsection{LLM-as-a-Judge Results}
\label{sec:llm-as-a-judge}

\begin{table}[t]
\centering
\small
\setlength{\tabcolsep}{3pt}
\renewcommand{\arraystretch}{1.1}

\begin{tabular}{lcc}
\toprule
\multirow{2.5}{*}{Model} & \multicolumn{2}{c}{\textbf{LLM-as-a-judge}} \\
\cmidrule(lr){2-3}
 & \texttt{en}$\rightarrow$\texttt{xx} & \texttt{xx}$\rightarrow$\texttt{en} \\
\midrule
Gemma2-9B
&  &  \\
\midrule
\quad+ DPO         & 84.30 & 92.59 \\
\quad\quad+ CLewR  & 83.71 & 92.49 \\
\hdashline
\quad+ CPO         & 85.40 & 92.08 \\
\quad\quad+ CLewR  & 87.99 & 93.99 \\
\hdashline
\quad+ ARPO        & 88.84 & 94.15 \\
\quad\quad+ CLewR  & 88.54 & 94.42 \\
\hdashline
\quad+ ARPO-$z'$-V1  & 90.23 & 94.76 \\
\quad\quad+ CLewR  & 90.00 & 95.14 \\
\hdashline
\quad+ ARPO-$z'$-V2  & \textbf{90.41} & 94.80 \\
\quad\quad+ CLewR  & 90.04 & \textbf{95.21} \\

% \midrule

% Qwen
% &  &  \\
% \quad ARPO        & 76.98 & 87.81 \\
% \quad\quad CLewR  & 75.87 & 90.15 \\

% \quad ARPO-z'-V1  & 77.52 & 89.48 \\
% \quad\quad CLewR  & 77.43 & 89.80 \\

% \quad ARPO-z'-V2  & 77.59 & 90.90 \\
% \quad\quad CLewR  & 77.42 & 90.14 \\

% \quad CPO         & 70.61 & 85.83 \\
% \quad\quad CLewR  & 71.27 & 88.35 \\

% \quad DPO         & 69.19 & 89.90 \\
% \quad\quad CLewR  & 68.79 & 89.47 \\
% LLaMA
% &  &  \\
% \quad ARPO        & 86.01 & 92.45 \\
% \quad\quad CLewR  & 85.39 & 92.37 \\

% \quad ARPO-z'-V1  & 86.04 & 92.90 \\
% \quad\quad CLewR  & 87.02 & 93.22 \\

% \quad ARPO-z'-V2  & 86.16 & 92.38 \\
% \quad\quad CLewR  & 86.90 & 93.13 \\

% \quad CPO         & 16.43 & 64.65 \\
% \quad\quad CLewR  & 22.71 & 59.82 \\

% \quad DPO         & 83.52 & 91.55 \\
% \quad\quad CLewR  & 83.41 & 91.85 \\

\bottomrule
\end{tabular}
\vspace{-0.2cm}
\caption{LLM-as-a-judge evaluation results for Gemma2-9B. Results are averaged after running on the three predictions for each PO method.}
\label{tab:llm_judge}
\vspace{-0.25cm}
\end{table}

In addition to automatic metrics, we evaluate translation quality using an LLM-as-a-judge setup, where a large language model assigns a scalar score to each translation based on adequacy and fluency. This complementary evaluation provides a more holistic assessment that is less dependent on surface-level overlap with reference translations.

We use GPT5.1 (reasoning effort ``medium'') to judge MT results,  with the following prompt:

\begin{tcolorbox}[
    colback=white,
    colframe=black,
    boxrule=0.2pt,
    breakable,
    width=\columnwidth
]
\texttt{JUDGE\_PROMPT\_TEMPLATE}

\small
Score the following translation from \{source\_lang\_name\} to \{target\_lang\_name\} on a scale from 0 to 100, where a score of 0 means a broken or poor translation; 33 indicates a flawed translation with significant issues; 66 indicates a good translation with only minor issues in grammar, fluency, or consistency; and 100 represents a perfect translation in both meaning and grammar. Answer with only a whole number.

% \medskip
\textbf{\{source\_lang\_name\} source text:} \{source\_text\}

\textbf{\{target\_lang\_name\} translation:} \{model\_output\}
\end{tcolorbox}

 The LLM-as-a-judge results are reported in Table~\ref{tab:llm_judge}. We observe that the GPT5.1 judge produces results that are mostly consistent with the usual BLEU and COMET performance metrics.

\subsection{Detailed Results and Borda Counts}
\label{sec:borda}

To provide a more comprehensive comparison across language pairs and translation directions, we report Borda count rankings \cite{emerson2013original} based on BLEU and COMET scores. The detailed results are presented in Tables~\ref{tab:borda_en_xx} and~\ref{tab:borda_xx_en}, which include per-language-pair scores, as well as aggregated rankings.

For each metric, systems are first ranked independently on each language pair and translation direction. We then compute the average rank across all language pairs, corresponding to a standard Borda count aggregation. We consider two variants of this procedure: (i) \emph{per model} (M), where all model configurations are jointly ranked for each language pair, and (ii) \emph{per method} (m), where each base method is ranked together with its corresponding variants, e.g.~CLewR and CLewR-$z$.

Since the number of systems $N$ differs between these settings, we linearly rescale the average rank to the interval $[0, 100]$, where $100$ corresponds to a consistently top rank, and $0$ to a consistently bottom rank across all language pairs. The rescaled score is computed as:
\begin{equation}
100 \cdot \left(1 - \frac{R - 1}{N - 1}\right),    
\end{equation}
where $R$ denotes the average rank. The last columns in Tables~\ref{tab:borda_en_xx} and~\ref{tab:borda_xx_en} report these normalized Borda scores for both BLEU and COMET, under the per-model (M) and per-method (m) settings, respectively.

Overall, the Borda rankings are consistent with the trends observed in the main experiments, confirming that CLewR-based variants generally achieve stronger aggregate performance across language pairs and directions. In particular, the improvements are reflected in both BLEU-based and COMET-based rankings under both per-model and per-method settings. This further supports the robustness of our conclusions beyond individual metric values.

% \begin{table*}[ht]
% \centering
% \scriptsize
% \setlength{\tabcolsep}{2pt}
% \begin{tabular}{@{}l|cc|cc|cc|cc|cc|cc|cc|cc@{}}
% \toprule
% Method & \multicolumn{2}{c}{en$\rightarrow$ca} & \multicolumn{2}{c}{en$\rightarrow$es} & \multicolumn{2}{c}{en$\rightarrow$gl} & \multicolumn{2}{c}{en$\rightarrow$it} & \multicolumn{2}{c}{en$\rightarrow$pt} & \multicolumn{2}{c}{en$\rightarrow$ro} & \multicolumn{4}{c}{Borda count} \\
%  & BLEU & COMET & BLEU & COMET & BLEU & COMET & BLEU & COMET & BLEU & COMET & BLEU & COMET & \rotatebox{90}{BLEU(M)} & \rotatebox{90}{COMET(M)} & \rotatebox{90}{BLEU(m)} & \rotatebox{90}{COMET(m)} \\
% \midrule

\subsection{Easy vs.~Hard Examples for Curriculum}
\label{sec:examples}

In Figure~\ref{fig:examples}, we list several ($x$, $y_{w}$, $y_{l}$) triplets that are either easy (low similarity $s(y_{w}$, $y_{l})$) or hard (high similarity $s(y_{w}$, $y_{l})$). For the hard examples, we underline the differences between chosen and rejected sentences.

\begin{table*}[ht]
\centering
\setlength{\tabcolsep}{2pt}
\renewcommand{\arraystretch}{1.2}
\resizebox{0.98\linewidth}{!}{
\begin{tabular}{@{}l|cc|cc|cc|cc|cc|cc|cc|cc@{}}
\toprule
\multicolumn{1}{l}{\multirow{3.5}{*}{Model}} 
& \multicolumn{2}{c}{\texttt{en}$\rightarrow$\texttt{ca}} 
& \multicolumn{2}{c}{\texttt{en}$\rightarrow$\texttt{es}} 
& \multicolumn{2}{c}{\texttt{en}$\rightarrow$\texttt{gl}} 
& \multicolumn{2}{c}{\texttt{en}$\rightarrow$\texttt{it}} 
& \multicolumn{2}{c}{\texttt{en}$\rightarrow$\texttt{pt}} 
& \multicolumn{2}{c}{\texttt{en}$\rightarrow$\texttt{ro}} 
& \multicolumn{2}{c}{Borda (M)} 
& \multicolumn{2}{c}{Borda (m)} \\
\cmidrule(lr){2-13} \cmidrule(lr){14-17}
& \multirow{2}{*}{\rotatebox[origin=c]{50}{BLEU}} 
& \multirow{2}{*}{\rotatebox[origin=c]{50}{COMET}} 
& \multirow{2}{*}{\rotatebox[origin=c]{50}{BLEU}} 
& \multirow{2}{*}{\rotatebox[origin=c]{50}{COMET}} 
& \multirow{2}{*}{\rotatebox[origin=c]{50}{BLEU}} 
& \multirow{2}{*}{\rotatebox[origin=c]{50}{COMET}} 
& \multirow{2}{*}{\rotatebox[origin=c]{50}{BLEU}} 
& \multirow{2}{*}{\rotatebox[origin=c]{50}{COMET}} 
& \multirow{2}{*}{\rotatebox[origin=c]{50}{BLEU}} 
& \multirow{2}{*}{\rotatebox[origin=c]{50}{COMET}} 
& \multirow{2}{*}{\rotatebox[origin=c]{50}{BLEU}} 
& \multirow{2}{*}{\rotatebox[origin=c]{50}{COMET}} 
& \multirow{2}{*}{\rotatebox[origin=c]{50}{BLEU}} 
& \multirow{2}{*}{\rotatebox[origin=c]{50}{COMET}} 
& \multirow{2}{*}{\rotatebox[origin=c]{50}{BLEU}} 
& \multirow{2}{*}{\rotatebox[origin=c]{50}{COMET}} \\
% & \multicolumn{2}{c|}{Model} 
% & \multicolumn{2}{c}{Method} \\
&  &  &  &  &  &  &  &  &  &  &  &  
& & & & \\
% & \rotatebox[origin=c]{50}{BLEU} & \rotatebox[origin=c]{50}{COMET} & \rotatebox[origin=c]{50}{BLEU} & \rotatebox[origin=c]{50}{COMET} \\
\midrule
Gemma2-9B & 12.35 & 56.90 & 15.62 & 57.20 & 18.57 & 62.96 & 19.15 & 64.46 & 37.66 & 76.35 & 15.74 & 56.63 & 7.14 & 0.00 &  &  \\
\midrule

\quad + DPO & 26.14 & 84.98 & 19.74 & 85.31 & 16.33 & 82.40 & 19.55 & 86.75 & 31.30 & 88.37 & 26.49 & 87.82 & 11.90 & 15.48 & 50.00 & 58.33 \\
\quad\quad + CurriDPO & 26.41 & 84.57 & 16.02 & 84.56 & 17.81 & 82.93 & 20.28 & 86.52 & 27.35 & 87.77 & 23.00 & 87.46 & 13.10 & 9.52 & 58.33 & 16.67 \\
\quad\quad + CLewR & 22.72 & 84.87 & 18.56 & 85.46 & 16.83 & 83.22 & 20.28 & 87.01 & 32.91 & 88.63 & 22.77 & 87.45 & 11.90 & 17.86 & 50.00 & 75.00 \\
\hdashline
\quad + CPO & 36.72 & 86.42 & 28.21 & 86.49 & 29.32 & 85.03 & 28.54 & 88.10 & 44.18 & 89.17 & 34.22 & 89.37 & 28.57 & 28.57 & 0.00 & 0.00 \\
\quad\quad + CLewR & 40.33 & 86.78 & 28.33 & 86.61 & 34.59 & 86.14 & 29.58 & 88.11 & 47.47 & 89.63 & 37.17 & 89.83 & 48.81 & 35.71 & 100.00 & 100.00 \\
\hdashline
\quad + ARPO & 38.56 & 86.93 & 28.70 & 86.97 & 33.28 & 86.61 & 29.51 & 88.61 & 46.09 & 89.74 & 36.10 & 89.86 & 38.10 & 52.38 & 8.33 & 50.00 \\
\quad\quad + CLewR & 40.59 & 87.21 & 28.64 & 86.83 & 35.02 & 86.52 & 29.91 & 88.33 & 47.96 & 89.84 & 37.66 & 89.97 & 61.90 & 51.19 & 58.33 & 41.67 \\
\quad\quad + CLewR-$z$ & 40.06 & 87.04 & 28.78 & 86.85 & 35.07 & 86.66 & 29.95 & 88.46 & 47.33 & 89.64 & 37.79 & 90.05 & 60.71 & 51.19 & 83.33 & 58.33 \\
\hdashline
\quad + ARPO-$z'$-V1 & 38.91 & 87.16 & 29.66 & 87.09 & 34.49 & 87.20 & 30.29 & 88.68 & 47.52 & 89.94 & 37.21 & 90.32 & 61.90 & 83.33 & 16.67 & 58.33 \\
\quad\quad + CLewR & 41.44 & 87.47 & 29.22 & 86.99 & 35.74 & 87.02 & 30.42 & 88.52 & 48.91 & 89.98 & 38.93 & 90.42 & 91.67 & 85.71 & 83.33 & 58.33 \\
\quad\quad + CLewR-$z$ & 40.80 & 87.28 & 28.73 & 86.99 & 35.20 & 87.17 & 30.43 & 88.72 & 48.49 & 89.89 & 38.22 & 90.24 & 78.57 & 78.57 & 50.00 & 41.67 \\
\hdashline
\quad + ARPO-$z'$-V2 & 39.64 & 87.22 & 29.56 & 87.09 & 34.41 & 87.28 & 30.15 & 88.67 & 47.69 & 89.97 & 37.32 & 90.32 & 61.90 & 89.29 & 16.67 & 75.00 \\
\quad\quad + CLewR & 41.45 & 87.34 & 29.14 & 87.01 & 35.72 & 86.97 & 30.40 & 88.50 & 48.97 & 89.97 & 38.99 & 90.42 & 91.67 & 82.14 & 83.33 & 58.33 \\
\quad\quad + CLewR-$z$ & 40.47 & 87.21 & 28.91 & 86.98 & 35.55 & 87.14 & 30.69 & 88.73 & 48.55 & 89.94 & 38.29 & 90.28 & 83.33 & 77.38 & 50.00 & 25.00 \\

\midrule
Qwen2.5-7B & 3.26 & 44.52 & 3.50 & 47.39 & 2.98 & 43.89 & 3.19 & 45.88 & 6.57 & 48.85 & 5.11 & 45.13 & 0.00 & 0.00 &  &  \\
\midrule

\quad + DPO & 23.80 & 78.95 & 24.36 & 85.29 & 17.19 & 80.00 & 22.67 & 85.94 & 36.17 & 87.96 & 22.41 & 79.25 & 16.67 & 21.43 & 66.67 & 91.67 \\
\quad\quad + CurriDPO & 24.99 & 78.19 & 22.14 & 84.29 & 18.53 & 78.72 & 22.35 & 85.25 & 32.30 & 87.10 & 22.50 & 77.83 & 16.67 & 7.14 & 66.67 & 0.00 \\
\quad\quad + CLewR & 23.52 & 78.92 & 21.51 & 84.64 & 18.08 & 80.36 & 21.88 & 85.60 & 34.52 & 87.71 & 22.02 & 79.21 & 9.52 & 15.48 & 16.67 & 58.33 \\
\hdashline
\quad + CPO & 28.07 & 80.46 & 25.85 & 85.61 & 24.32 & 81.95 & 23.75 & 85.97 & 38.65 & 88.11 & 25.42 & 81.46 & 29.76 & 30.95 & 16.67 & 33.33 \\
\quad\quad + CLewR & 31.51 & 81.15 & 25.80 & 85.31 & 27.37 & 82.49 & 25.14 & 85.70 & 42.01 & 88.28 & 28.50 & 81.96 & 35.71 & 32.14 & 83.33 & 66.67 \\
\hdashline
\quad + ARPO & 31.58 & 83.26 & 26.87 & 86.36 & 28.41 & 84.46 & 25.20 & 86.82 & 41.49 & 88.74 & 28.93 & 84.17 & 46.43 & 70.24 & 16.67 & 83.33 \\
\quad\quad + CLewR & 33.76 & 82.81 & 26.53 & 85.99 & 29.14 & 83.84 & 25.99 & 86.76 & 43.61 & 88.79 & 30.33 & 84.07 & 65.48 & 46.43 & 75.00 & 25.00 \\
\quad\quad + CLewR-$z$ & 32.89 & 82.87 & 26.67 & 86.07 & 29.28 & 84.28 & 25.63 & 86.71 & 42.41 & 88.67 & 29.95 & 84.18 & 58.33 & 53.57 & 58.33 & 41.67 \\
\hdashline
\quad + ARPO-$z'$-V1 & 32.94 & 83.04 & 27.60 & 86.40 & 28.64 & 84.26 & 25.68 & 86.98 & 42.21 & 88.90 & 29.37 & 84.46 & 64.29 & 75.00 & 16.67 & 33.33 \\
\quad\quad + CLewR & 34.78 & 83.43 & 27.41 & 86.20 & 30.17 & 84.15 & 26.84 & 86.89 & 44.91 & 89.10 & 31.94 & 85.11 & 89.29 & 76.19 & 75.00 & 33.33 \\
\quad\quad + CLewR-$z$ & 34.61 & 83.64 & 27.18 & 86.34 & 30.61 & 84.35 & 26.72 & 87.11 & 44.81 & 89.11 & 32.16 & 85.01 & 84.52 & 91.67 & 58.33 & 83.33 \\
\hdashline
\quad + ARPO-$z'$-V2 & 32.69 & 83.30 & 27.48 & 86.28 & 28.27 & 84.20 & 25.86 & 86.99 & 42.52 & 88.88 & 28.87 & 84.37 & 59.52 & 70.24 & 16.67 & 25.00 \\
\quad\quad + CLewR & 34.99 & 83.52 & 27.42 & 86.21 & 30.12 & 84.10 & 26.76 & 86.78 & 44.85 & 89.03 & 32.09 & 85.07 & 89.29 & 71.43 & 75.00 & 33.33 \\
\quad\quad + CLewR-$z$ & 34.46 & 83.62 & 27.22 & 86.33 & 30.78 & 84.36 & 26.70 & 87.13 & 44.86 & 89.10 & 31.99 & 85.00 & 84.52 & 89.29 & 58.33 & 91.67 \\

\midrule
LLaMA3.1-8B & 1.95 & 42.78 & 1.34 & 46.25 & 1.49 & 45.12 & 1.42 & 47.07 & 2.18 & 48.00 & 1.82 & 46.75 & 7.14 & 0.00 &  &  \\
\midrule

\quad + DPO & 34.06 & 86.08 & 27.28 & 85.83 & 27.16 & 85.08 & 27.36 & 87.66 & 40.96 & 88.76 & 32.33 & 88.87 & 28.57 & 36.90 & 50.00 & 83.33 \\
\quad\quad + CurriDPO & 35.92 & 86.10 & 26.71 & 85.80 & 30.06 & 85.18 & 27.62 & 87.43 & 40.98 & 88.59 & 33.47 & 88.56 & 38.10 & 28.57 & 91.67 & 50.00 \\
\quad\quad + CLewR & 33.36 & 85.93 & 26.48 & 85.62 & 27.82 & 85.03 & 27.04 & 87.46 & 39.02 & 88.45 & 31.51 & 88.57 & 22.62 & 23.81 & 8.33 & 16.67 \\
\hdashline
\quad + CPO & 1.11 & 52.39 & 1.23 & 60.19 & 1.31 & 50.18 & 2.32 & 61.93 & 1.63 & 56.53 & 1.23 & 48.62 & 1.19 & 8.33 & 0.00 & 16.67 \\
\quad\quad + CLewR & 2.70 & 48.81 & 4.99 & 66.26 & 1.42 & 56.03 & 4.77 & 65.96 & 7.59 & 71.15 & 2.33 & 49.78 & 13.10 & 13.10 & 100.00 & 83.33 \\
\hdashline
\quad + ARPO & 36.61 & 86.40 & 27.45 & 86.46 & 32.68 & 86.30 & 27.62 & 87.82 & 43.15 & 89.03 & 33.47 & 88.83 & 47.62 & 55.95 & 16.67 & 58.33 \\
\quad\quad + CLewR & 37.90 & 86.40 & 27.35 & 86.37 & 33.30 & 86.12 & 28.05 & 87.69 & 45.17 & 89.28 & 35.16 & 88.80 & 64.29 & 51.19 & 66.67 & 25.00 \\
\quad\quad + CLewR-$z$ & 37.72 & 86.60 & 27.49 & 86.55 & 33.36 & 86.26 & 27.53 & 87.81 & 44.53 & 89.13 & 35.32 & 89.07 & 60.71 & 67.86 & 66.67 & 75.00 \\
\hdashline
\quad + ARPO-$z'$-V1 & 36.81 & 86.36 & 28.19 & 86.54 & 31.99 & 86.06 & 27.92 & 87.84 & 43.31 & 89.05 & 33.84 & 88.78 & 57.14 & 53.57 & 16.67 & 8.33 \\
\quad\quad + CLewR & 38.95 & 86.86 & 27.90 & 86.53 & 34.80 & 86.49 & 28.52 & 88.00 & 46.16 & 89.38 & 36.64 & 89.34 & 86.90 & 86.90 & 66.67 & 58.33 \\
\quad\quad + CLewR-$z$ & 39.08 & 86.97 & 27.91 & 86.63 & 34.33 & 86.28 & 28.53 & 88.02 & 45.59 & 89.40 & 36.56 & 89.34 & 84.52 & 90.48 & 66.67 & 91.67 \\
\hdashline
\quad + ARPO-$z'$-V2 & 37.18 & 86.45 & 28.43 & 86.50 & 32.54 & 86.15 & 28.00 & 87.84 & 43.64 & 89.07 & 33.99 & 88.93 & 64.29 & 61.90 & 16.67 & 0.00 \\
\quad\quad + CLewR & 39.02 & 86.98 & 27.94 & 86.52 & 34.89 & 86.48 & 28.67 & 87.86 & 46.12 & 89.48 & 36.59 & 89.32 & 91.67 & 85.71 & 75.00 & 75.00 \\
\quad\quad + CLewR-$z$ & 39.31 & 87.05 & 28.08 & 86.70 & 34.29 & 86.37 & 28.50 & 87.98 & 45.89 & 89.40 & 36.49 & 89.27 & 84.52 & 90.48 & 58.33 & 75.00 \\
\bottomrule
\end{tabular}
}
\caption{Results across several base models and preference optimization methods, with and without CLewR, for each pair of languages of the form \texttt{en}$\rightarrow$\texttt{xx}. Normalized Borda counts are computed per model (M) and per method (m), respectively.}
\label{tab:borda_en_xx}
\end{table*}

% \begin{table*}[ht]
% \centering
% \scriptsize
% \setlength{\tabcolsep}{2pt}
% \begin{tabular}{@{}l|c|c|c|c|c|c|c|c|c|c|c|c|c|c|c|c@{}}
% \toprule
% Method & \multicolumn{2}{c}{ca$\rightarrow$en} & \multicolumn{2}{c}{es$\rightarrow$en} & \multicolumn{2}{c}{gl$\rightarrow$en} & \multicolumn{2}{c}{it$\rightarrow$en} & \multicolumn{2}{c}{pt$\rightarrow$en} & \multicolumn{2}{c}{ro$\rightarrow$en} & \multicolumn{4}{c}{Borda count} \\
%  & BLEU & COMET & BLEU & COMET & BLEU & COMET & BLEU & COMET & BLEU & COMET & BLEU & COMET & \rotatebox{90}{BLEU(m)} & \rotatebox{90}{COMET(m)} & \rotatebox{90}{BLEU(M)} & \rotatebox{90}{COMET(M)} \\
% \midrule

\begin{table*}[ht]
\centering
\setlength{\tabcolsep}{2pt}
\renewcommand{\arraystretch}{1.2}
\resizebox{0.98\linewidth}{!}{
\begin{tabular}{@{}l|cc|cc|cc|cc|cc|cc|cc|cc@{}}
\toprule
\multicolumn{1}{l}{\multirow{3.5}{*}{Model}} 
& \multicolumn{2}{c}{\texttt{ca}$\rightarrow$\texttt{en}} 
& \multicolumn{2}{c}{\texttt{es}$\rightarrow$\texttt{en}} 
& \multicolumn{2}{c}{\texttt{gl}$\rightarrow$\texttt{en}} 
& \multicolumn{2}{c}{\texttt{it}$\rightarrow$\texttt{en}} 
& \multicolumn{2}{c}{\texttt{pt}$\rightarrow$\texttt{en}} 
& \multicolumn{2}{c}{\texttt{ro}$\rightarrow$\texttt{en}} 
& \multicolumn{2}{c}{Borda (M)} 
& \multicolumn{2}{c}{Borda (m)} \\
\cmidrule(lr){2-13} \cmidrule(lr){14-17}
& \multirow{2}{*}{\rotatebox[origin=c]{50}{BLEU}} 
& \multirow{2}{*}{\rotatebox[origin=c]{50}{COMET}} 
& \multirow{2}{*}{\rotatebox[origin=c]{50}{BLEU}} 
& \multirow{2}{*}{\rotatebox[origin=c]{50}{COMET}} 
& \multirow{2}{*}{\rotatebox[origin=c]{50}{BLEU}} 
& \multirow{2}{*}{\rotatebox[origin=c]{50}{COMET}} 
& \multirow{2}{*}{\rotatebox[origin=c]{50}{BLEU}} 
& \multirow{2}{*}{\rotatebox[origin=c]{50}{COMET}} 
& \multirow{2}{*}{\rotatebox[origin=c]{50}{BLEU}} 
& \multirow{2}{*}{\rotatebox[origin=c]{50}{COMET}} 
& \multirow{2}{*}{\rotatebox[origin=c]{50}{BLEU}} 
& \multirow{2}{*}{\rotatebox[origin=c]{50}{COMET}} 
& \multirow{2}{*}{\rotatebox[origin=c]{50}{BLEU}} 
& \multirow{2}{*}{\rotatebox[origin=c]{50}{COMET}} 
& \multirow{2}{*}{\rotatebox[origin=c]{50}{BLEU}} 
& \multirow{2}{*}{\rotatebox[origin=c]{50}{COMET}} \\
% & \multicolumn{2}{c|}{Model} 
% & \multicolumn{2}{c}{Method} \\
&  &  &  &  &  &  &  &  &  &  &  &  
& & & & \\
% & \rotatebox[origin=c]{50}{BLEU} & \rotatebox[origin=c]{50}{COMET} & \rotatebox[origin=c]{50}{BLEU} & \rotatebox[origin=c]{50}{COMET} \\
\midrule
Gemma2-9B & 37.48 & 84.90 & 26.41 & 83.38 & 34.35 & 85.40 & 30.35 & 86.63 & 44.79 & 87.32 & 36.52 & 85.79 & 21.43 & 0.00 &  &  \\
\midrule

\quad + DPO & 38.82 & 88.08 & 28.61 & 87.12 & 25.88 & 86.19 & 29.56 & 87.66 & 41.29 & 88.89 & 34.84 & 88.67 & 15.48 & 20.24 & 91.67 & 83.33 \\
\quad\quad + CurriDPO & 33.84 & 87.60 & 25.10 & 86.73 & 22.57 & 85.62 & 27.14 & 87.34 & 32.52 & 88.17 & 28.84 & 88.18 & 1.19 & 8.33 & 8.33 & 0.00 \\
\quad\quad + CLewR & 37.35 & 88.01 & 28.07 & 87.16 & 22.54 & 85.63 & 30.43 & 87.70 & 38.14 & 88.74 & 34.63 & 88.66 & 9.52 & 17.86 & 50.00 & 66.67 \\
\hdashline
\quad + CPO & 41.61 & 88.17 & 30.74 & 87.17 & 29.92 & 85.57 & 32.70 & 87.82 & 43.95 & 88.97 & 39.70 & 88.86 & 26.19 & 25.00 & 0.00 & 0.00 \\
\quad\quad + CLewR & 45.53 & 88.82 & 32.76 & 87.63 & 36.95 & 87.15 & 35.03 & 88.31 & 48.24 & 89.58 & 43.52 & 89.46 & 59.52 & 48.81 & 100.00 & 100.00 \\
\hdashline
\quad + ARPO & 42.58 & 88.60 & 31.08 & 87.49 & 33.72 & 86.95 & 33.28 & 88.12 & 44.77 & 89.34 & 40.64 & 89.32 & 33.33 & 35.71 & 0.00 & 0.00 \\
\quad\quad + CLewR & 45.55 & 88.84 & 32.95 & 87.67 & 37.17 & 87.42 & 35.02 & 88.35 & 48.48 & 89.61 & 43.46 & 89.49 & 61.90 & 59.52 & 91.67 & 83.33 \\
\quad\quad + CLewR-$z$ & 44.16 & 88.74 & 32.53 & 87.70 & 38.93 & 88.03 & 34.32 & 88.27 & 47.36 & 89.56 & 41.95 & 89.43 & 47.62 & 51.19 & 58.33 & 66.67 \\
\hdashline
\quad + ARPO-$z'$-V1 & 44.86 & 88.83 & 33.23 & 87.72 & 37.58 & 87.71 & 35.42 & 88.38 & 47.77 & 89.59 & 43.10 & 89.49 & 60.71 & 65.48 & 0.00 & 0.00 \\
\quad\quad + CLewR & 45.98 & 88.97 & 33.45 & 87.78 & 39.14 & 88.05 & 35.67 & 88.47 & 49.07 & 89.74 & 44.36 & 89.66 & 90.48 & 86.90 & 75.00 & 58.33 \\
\quad\quad + CLewR-$z$ & 45.84 & 89.04 & 33.80 & 87.82 & 39.59 & 88.31 & 35.70 & 88.50 & 48.77 & 89.76 & 43.64 & 89.64 & 85.71 & 95.24 & 75.00 & 91.67 \\
\hdashline
\quad + ARPO-$z'$-V2 & 45.00 & 88.86 & 33.24 & 87.71 & 36.36 & 87.44 & 35.26 & 88.38 & 47.65 & 89.53 & 43.10 & 89.54 & 57.14 & 63.10 & 0.00 & 0.00 \\
\quad\quad + CLewR & 45.93 & 88.93 & 33.47 & 87.79 & 39.03 & 88.05 & 35.48 & 88.44 & 48.92 & 89.72 & 43.70 & 89.59 & 83.33 & 80.95 & 50.00 & 50.00 \\
\quad\quad + CLewR-$z$ & 46.04 & 89.03 & 33.92 & 87.87 & 40.32 & 88.40 & 35.72 & 88.50 & 48.96 & 89.78 & 43.74 & 89.64 & 97.62 & 97.62 & 100.00 & 100.00 \\

\midrule
Qwen2.5-7B & 10.50 & 63.35 & 7.67 & 62.83 & 10.90 & 64.15 & 8.30 & 62.67 & 11.54 & 63.22 & 8.66 & 60.76 & 0.00 & 0.00 &  &  \\
\midrule

\quad + DPO & 36.38 & 87.33 & 28.19 & 87.03 & 28.82 & 86.17 & 29.62 & 87.54 & 40.58 & 88.74 & 34.73 & 87.71 & 35.71 & 53.57 & 100.00 & 100.00 \\
\quad\quad + CurriDPO & 30.94 & 86.60 & 25.34 & 86.47 & 26.04 & 85.43 & 26.44 & 87.03 & 35.83 & 88.12 & 30.14 & 87.00 & 8.33 & 20.24 & 8.33 & 8.33 \\
\quad\quad + CLewR & 34.85 & 87.21 & 26.38 & 86.78 & 22.51 & 85.29 & 28.68 & 87.32 & 38.37 & 88.48 & 33.02 & 87.51 & 17.86 & 27.38 & 41.67 & 41.67 \\
\hdashline
\quad + CPO & 35.29 & 86.38 & 27.77 & 86.12 & 27.04 & 84.43 & 28.65 & 86.61 & 37.99 & 87.69 & 33.20 & 86.93 & 20.24 & 7.14 & 0.00 & 0.00 \\
\quad\quad + CLewR & 39.69 & 87.24 & 30.82 & 86.91 & 33.37 & 85.95 & 32.04 & 87.45 & 43.80 & 88.56 & 37.95 & 87.84 & 70.24 & 40.48 & 100.00 & 100.00 \\
\hdashline
\quad + ARPO & 35.81 & 86.76 & 27.92 & 86.41 & 27.87 & 84.82 & 28.59 & 86.92 & 38.01 & 87.95 & 34.51 & 87.41 & 25.00 & 16.67 & 0.00 & 0.00 \\
\quad\quad + CLewR & 40.56 & 87.62 & 31.61 & 87.22 & 32.72 & 86.24 & 32.47 & 87.65 & 44.12 & 88.94 & 38.70 & 88.04 & 79.76 & 84.52 & 100.00 & 100.00 \\
\quad\quad + CLewR-$z$ & 38.99 & 87.49 & 29.79 & 87.08 & 31.97 & 86.07 & 30.49 & 87.52 & 41.68 & 88.62 & 36.37 & 87.92 & 44.05 & 57.14 & 50.00 & 50.00 \\
\hdashline
\quad + ARPO-$z'$-V1 & 38.95 & 87.42 & 30.25 & 86.99 & 32.01 & 86.00 & 30.70 & 87.38 & 41.99 & 88.56 & 36.84 & 87.89 & 48.81 & 46.43 & 0.00 & 8.33 \\
\quad\quad + CLewR & 41.29 & 87.61 & 31.69 & 87.10 & 33.43 & 86.22 & 32.86 & 87.66 & 44.22 & 88.86 & 39.55 & 88.13 & 95.24 & 80.95 & 100.00 & 75.00 \\
\quad\quad + CLewR-$z$ & 40.94 & 87.71 & 31.18 & 87.22 & 32.91 & 85.97 & 31.91 & 87.51 & 44.02 & 88.87 & 38.30 & 88.11 & 76.19 & 76.19 & 50.00 & 66.67 \\
\hdashline
\quad + ARPO-$z'$-V2 & 40.11 & 87.78 & 31.19 & 87.27 & 34.43 & 86.77 & 31.41 & 87.68 & 43.39 & 88.85 & 37.88 & 88.12 & 69.05 & 91.67 & 25.00 & 75.00 \\
\quad\quad + CLewR & 41.37 & 87.71 & 31.74 & 87.17 & 33.17 & 86.21 & 33.05 & 87.74 & 44.15 & 88.86 & 39.50 & 88.16 & 94.05 & 88.10 & 91.67 & 66.67 \\
\quad\quad + CLewR-$z$ & 40.74 & 87.59 & 30.67 & 87.21 & 32.13 & 85.76 & 31.75 & 87.58 & 43.61 & 88.80 & 37.99 & 88.10 & 65.48 & 64.29 & 33.33 & 8.33 \\

\midrule
LLaMA3.1-8B & 2.11 & 44.96 & 1.43 & 45.01 & 1.81 & 44.15 & 1.57 & 47.36 & 2.12 & 47.70 & 2.03 & 48.65 & 0.00 & 0.00 &  &  \\
\midrule

\quad + DPO & 40.90 & 88.20 & 30.22 & 87.18 & 32.05 & 86.32 & 31.73 & 87.83 & 42.21 & 88.73 & 37.54 & 88.64 & 32.14 & 46.43 & 50.00 & 58.33 \\
\quad\quad + CurriDPO & 40.88 & 87.99 & 29.92 & 87.09 & 33.36 & 86.51 & 30.52 & 87.68 & 43.46 & 88.66 & 37.92 & 88.56 & 34.52 & 28.57 & 58.33 & 16.67 \\
\quad\quad + CLewR & 39.33 & 88.17 & 30.19 & 87.19 & 31.25 & 86.25 & 31.93 & 87.84 & 42.46 & 88.83 & 37.60 & 88.68 & 30.95 & 51.19 & 41.67 & 75.00 \\
\hdashline
\quad + CPO & 13.01 & 74.51 & 10.33 & 75.97 & 6.30 & 70.18 & 10.74 & 76.56 & 13.03 & 73.82 & 8.73 & 74.85 & 7.14 & 7.14 & 0.00 & 0.00 \\
\quad\quad + CLewR & 15.39 & 77.11 & 11.54 & 77.73 & 10.90 & 72.15 & 12.27 & 77.68 & 15.14 & 77.67 & 13.19 & 78.24 & 14.29 & 14.29 & 100.00 & 100.00 \\
\hdashline
\quad + ARPO & 39.72 & 87.95 & 30.06 & 86.99 & 33.93 & 86.75 & 31.44 & 87.57 & 42.43 & 88.68 & 37.94 & 88.57 & 34.52 & 29.76 & 0.00 & 8.33 \\
\quad\quad + CLewR & 42.50 & 88.26 & 31.58 & 87.24 & 34.70 & 86.62 & 33.20 & 87.77 & 45.54 & 88.97 & 40.42 & 88.82 & 78.57 & 65.48 & 100.00 & 83.33 \\
\quad\quad + CLewR-$z$ & 40.98 & 88.22 & 30.69 & 87.20 & 34.30 & 86.78 & 32.21 & 87.72 & 43.43 & 88.86 & 38.85 & 88.78 & 50.00 & 58.33 & 50.00 & 58.33 \\
\hdashline
\quad + ARPO-$z'$-V1 & 42.01 & 88.13 & 31.60 & 87.17 & 34.73 & 86.89 & 33.16 & 87.74 & 44.60 & 88.77 & 40.85 & 88.75 & 71.43 & 48.81 & 16.67 & 0.00 \\
\quad\quad + CLewR & 43.67 & 88.54 & 32.06 & 87.38 & 36.34 & 87.29 & 34.18 & 87.94 & 46.33 & 89.17 & 41.71 & 89.01 & 98.81 & 96.43 & 100.00 & 91.67 \\
\quad\quad + CLewR-$z$ & 42.06 & 88.38 & 31.37 & 87.32 & 35.55 & 87.08 & 32.88 & 87.93 & 45.41 & 89.10 & 40.86 & 89.09 & 72.62 & 85.71 & 33.33 & 58.33 \\
\hdashline
\quad + ARPO-$z'$-V2 & 42.12 & 88.15 & 31.45 & 87.16 & 33.14 & 86.36 & 33.06 & 87.73 & 44.65 & 88.81 & 40.41 & 88.82 & 63.10 & 46.43 & 25.00 & 0.00 \\
\quad\quad + CLewR & 43.30 & 88.52 & 32.42 & 87.34 & 35.61 & 87.11 & 34.16 & 87.94 & 46.18 & 89.13 & 41.66 & 89.01 & 92.86 & 91.67 & 91.67 & 83.33 \\
\quad\quad + CLewR-$z$ & 42.21 & 88.39 & 31.07 & 87.25 & 36.07 & 87.33 & 32.68 & 87.81 & 44.89 & 88.99 & 40.28 & 89.02 & 69.05 & 83.33 & 33.33 & 66.67 \\
\bottomrule
\end{tabular}
}
\caption{Results across several base models and preference optimization methods, with and without CLewR, for each pair of languages of the form \texttt{xx}$\rightarrow$\texttt{en}. Normalized Borda counts are computed per model (M) and per method (m), respectively.}
\label{tab:borda_xx_en}
\end{table*}

% DIVERSITY METRICS

% \begin{table}[t]
% \centering
% \caption{Diversity (MATTR) and performance (TER, chrF) metrics.}
% \label{tab:a1}
% \begin{tabular}{lcccccc}
% \toprule
%  & \multicolumn{2}{c}{MATTR} & \multicolumn{2}{c}{TER} & \multicolumn{2}{c}{chrF} \\
% \cmidrule(lr){2-3} \cmidrule(lr){4-5} \cmidrule(lr){6-7}
% Method & en-xx & xx-en & en-xx & xx-en & en-xx & xx-en \\
% \midrule
% Gemma2-9B        & 85.87 & 83.15 & 91.29 & 61.31 & 39.11 & 65.66 \\
% \midrule
% \quad +DPO             & 69.10 & 78.97 & 102.90 & 61.82 & 54.48 & 62.62 \\
% \quad\quad  +curriDPO  & 67.70 & 71.92 & 107.03 & 82.23 & 54.44 & 60.56 \\
% \quad\quad  +CLewR     & 68.62 & 76.96 & 103.15 & 68.61 & 54.38 & 61.98 \\
% +CPO             & 83.15 & 82.64 & 54.36 & 51.20 & 60.07 & 63.04 \\
% \quad\quad  +CLewR     & 83.14 & 82.60 & 50.00 & 46.09 & 62.21 & 65.61 \\
% \quad +ARPO            & 83.06 & 82.75 & 51.02 & 49.18 & 61.49 & 64.16 \\
% \quad \quad +CLewR     & 83.33 & 82.65 & 49.43 & 45.88 & 62.50 & 65.73 \\
% \quad \quad +CLewR-z   & 83.53 & 82.75 & 49.81 & 46.74 & 62.41 & 65.74 \\
% \quad +ARPO-z'-V1      & 82.41 & 82.34 & 50.59 & 46.37 & 62.25 & 65.71 \\
% \quad \quad +CLewR     & 83.18 & 82.79 & 48.70 & 45.03 & 63.00 & 66.43 \\
% \quad \quad +CLewR-z   & 82.68 & 82.46 & 49.52 & 45.38 & 62.67 & 66.47 \\
% \quad +ARPO-z'-V2      & 82.56 & 82.41 & 50.34 & 46.78 & 62.37 & 65.54 \\
% \quad \quad +CLewR     & 83.18 & 82.82 & 48.67 & 45.21 & 63.05 & 66.42 \\
% \quad \quad +CLewR-z   & 82.77 & 82.38 & 49.34 & 45.18 & 62.74 & 66.61 \\
% \bottomrule
% \end{tabular}
% \end{table}

\end{document}